\documentclass{article}

\usepackage{amsmath}
\usepackage{amsfonts}
\usepackage{amssymb}
\usepackage{amsthm}
\usepackage{attrib}
\usepackage{booktabs}
\usepackage{calc}
\usepackage{caption}
\usepackage{changepage}
\usepackage{enumitem}
\usepackage{epstopdf} 
\usepackage{etoolbox}
\usepackage{fancyhdr}
\usepackage{filecontents}
\usepackage{float}
\usepackage{fontenc}
\usepackage{footmisc}
\usepackage{geometry}
\usepackage{graphicx}
\usepackage{hyperref} 
\usepackage{hyphenat} 
\usepackage{ifthen} 
\usepackage{inputenc}
\usepackage{indentfirst}
\usepackage{lastpage} 
\usepackage{lineno} 
\usepackage{mathpazo}
\usepackage{mathtools}
\usepackage{microtype}
\usepackage{multirow}
\usepackage[numbers]{natbib}
\usepackage{newfloat} 
\usepackage{pgfplots}
\usepackage{pgfplotstable}
\usepackage{setspace} 
\usepackage{soul}
\usepackage{subfigure}
\usepackage{tabto}
\usepackage{tikz} 
\usepackage{titlesec} 
\usepackage{totcount}
\usepackage{upgreek}
\usepackage{url}
\usepackage{verbatim}
\usepackage{xcolor}

\usepackage[range-phrase={\:to\:}]{siunitx}

\usetikzlibrary{angles, quotes}

\usepgfplotslibrary{statistics}

\DeclareMathOperator*{\argmin}{arg\,min}

\DeclarePairedDelimiter\abs{\lvert}{\rvert}%

\providecommand{\keywords}[1]
{
  \small	
  \textbf{\textit{Keywords---}} #1
}

\title{Pose Normalization of Indoor Mapping Datasets Partially Compliant to the Manhattan World Assumption}
\author{
    Patrick Hübner$^{1}$, Martin Weinmann$^{1}$, Sven Wursthorn$^{1}$, Stefan Hinz$^{1}$  \\
    \small $^{1}$Institute of Photogrammetry and Remote Sensing, Karlsruhe Institute of Technology, Germany
}

\date{}

\begin{document}
    \maketitle
    \begin{abstract}
        Due to their potential for a variety of applications, digital building models are meanwhile well established in planning, construction and usage phases of modern building projects.
        Older stock buildings however frequently lack digital representations and creating these manually is a tedious and time-consuming endeavor.
        For this reason, as well as due to the increasing availability of mobile indoor mapping systems facilitating the straight-forward, accurate and time-efficient capture of building structures, the automated reconstruction of building models from indoor mapping data has arisen as an active field of research.
        
        In this context, many approaches rely on simplifying suppositions about the structure of buildings to be reconstructed like e.g. the well-known Manhattan World assumption.
        This however not only presupposes that a given building structure itself is compliant to this assumption but also that the respective indoor mapping dataset is aligned accordingly with the coordinate axes.
        Indoor mapping systems on the other hand typically initialize the coordinate system arbitrarily by the sensor pose at the beginning of the mapping process.
        Indoor mapping data thus need to be aligned with the coordinate axes as a necessary preprocessing step for many indoor reconstruction approaches, which is also frequently known as pose normalization.
        Even in the case of indoor reconstruction approaches that do not rely on the Manhattan World assumption, such an alignment can be beneficial as it prevents aliasing effects when using data structures like voxel grids or octrees.
        Furthermore, it can be useful in the context of coregistering multiple indoor mapping datasets, for the automated analysis of architectural structures as well as for stabilization and drift-reduction in indoor mapping applications.
        
        In this paper, we present a novel pose normalization method for indoor mapping point clouds and triangle meshes that is robust against large fractions of the indoor mapping geometries deviating from an ideal Manhattan World structure.
        In the case of building structures that contain multiple Manhattan World systems, the dominant Manhattan World structure supported by the largest fraction of geometries is determined and used for alignment.
        In a first step, a vertical alignment orienting a chosen axis to be orthogonal to horizontal floor and ceiling surfaces is conducted.
        Subsequently, a rotation around the resulting vertical axis is determined that aligns the dataset horizontally with the coordinate axes.
        The proposed method is evaluated quantitatively against several publicly available indoor mapping datasets.
        Our implementation of the proposed procedure along with code for reproducing the evaluation will be made available to the public upon acceptance for publication.
    \end{abstract} 
    \hspace{10pt}
    \keywords{pose normalization, manhattan world, indoor mapping, point cloud, triangle mesh}
    \section{Introduction}
\label{sec:introduction}

The importance of digital models of building environments has been steadily increasing in recent years \citep{volk_et_al_2014, jung_lee_2015}.
Nowadays, many building projects are planned digitally in 3D using Building Information Modelling (BIM) techniques \citep{borrmann_et_al_2018}.
Thus, a valid digital three-dimensional model arises along with the construction of the respective building which can be profitably used during all the stages of the life cycle of a facility, i.e. usage and maintenance e.g. in the context of facility management, changes and modifications on the building and eventually dismantling \citep{arayici_et_al_2012, becker_et_al_2018,mirarchi_et_al_2018,gao_pishdad-bozorgi_2019}.
However, in the case of older, already existing buildings, three dimensional digital models often do not exist and two dimensional plans are often faulty or outdated.
Manually reconstructing digital models (as-is BIM models) for suchlike buildings is a tedious and time consuming process \citep{patraucean_et_al_2015, becker_et_al_2019}.

On the other hand, there currently exists a broad range of sensor systems that can be deployed to the task of accurately mapping indoor environments \citep{Lethola_et_al_2017,Chen_et_al_2018,Nocerino_et_al_2017,Masiero_et_al_2018}.
Terrestrial Laser Scanners (TLS) for instance can provide a high geometric accuracy of acquisition depending on the respective conditions e.g. in terms of surface characteristics and scanning geometry \citep{Soudarissanane_et_al_2011,Weinmann_2016}.
In order to achieve a complete capture of an environment however, multiple scans have to be conducted from varying positions.
Especially in the case of mapping the interior of building structures, this can be quite cumbersome as the device needs to be set up at numerous positions while the resulting scans subsequently need to be aligned.

Mobile mapping systems on the other hand alleviate these restrictions by continuously tracking their own position and orientation with respect to an initial pose.
Indoor mapping geometries acquired over time can thus be projected successively into a common coordinate system while the operator can achieve a complete scene capture by walking through the scene.
Mobile mapping systems encompass e.g. trolley-based (like NavVis\footnote{\url{https://www.navvis.com/m6}}) or backpack-mounted sensors \citep{Nuechter_et_al_2015,Blaser_et_al_2018,Wang_et_al_2018,karam_et_al_2019} or even UAV-based systems \citep{Hillemann_et_al_2019} as well as hand-carried (e.g. Leica BLK2GO\footnote{\url{https://blk2go.com}}) or head-worn devices (e.g. Microsoft HoloLens\footnote{\url{https://www.microsoft.com/de-de/hololens}}).
The latter, actually being an Augmented Reality (AR) system, offers the additional advantage of directly visualizing the already captured geometries within the view of the operator, facilitating the complete coverage of an indoor environment.

While conventional TLS or mobile laser scanning systems provide indoor mapping data in the form of point clouds, some consumer grade system like e.g. the mentioned Microsoft HoloLens or the Matterport system\footnote{\url{https://matterport.com/}} for instance, sometimes provide indoor mapping data in the form of preprocessed, condensed triangle meshes.
Such triangle meshes being a derived product from the primary point-based measurements were found to still provide sufficient accuracy for a wide range of applications \citep{bassier_et_al_2020b,weinmann_et_al_2020} while being significantly more compact in terms of data size and thus more efficient in terms of required processing time.

This broad range of available indoor mapping systems can provide an ample data base for the digital, three-dimensional reconstruction of built indoor environments.
Instead of having to take individual distance measurements in the respective building or having to bridge the mental gap between conventional, two-dimensional floor plans and the three-dimensional modeling environment, indoor mapping data representing existent buildings can be loaded directly into the modeling environment.
However, the manual digital reconstruction on the basis of indoor mapping data can still be a time-consuming endeavor.
Hence, automating this process has become the focus of a currently quite active field of research \citep{ma_liu_2018, kang_et_al_2020, pintore_et_al_2020}.

While recent approaches in the field of automated indoor reconstruction are becoming more flexible regarding the building structure represented by the indoor mapping data \citep{ochmann_et_al_2019,yang_et_al_2019,nikoohemat_et_al_2020b,tran_khoshelham_2020,wu_et_al_2020b, huebner_et_al_2021}, restricting assumptions about the building structure are still oftentimes applied.
A frequently applied simplification in this context is the Manhattan World assumption which is for instance relied upon in the indoor reconstruction approaches presented in \citep{furukawa_et_al_2009,gankhuyag_han_2020,otero_et_al_2020,shi_et_al_2020}.

The Manhattan World assumption, as first proposed by \citet{coughlan_yuille_1999,coughlan_yuille_2003}, presupposes all surfaces to be perpendicular to one of the three coordinate axes.
Applied to the context of building structures, this assumption thus prohibits curved room surfaces as well as surfaces being oriented diagonally with respect to the main building structure, i.e. diagonal walls or slanted ceilings.
The Manhattan World assumption has later been extended to the Atlanta World assumption by \citet{schindler_dellaert_2004} that weakens the Manhattan world assumption by permitting vertical surfaces to have arbitrary angles around a common vertical coordinate axis while horizontal surfaces are still expected to be perpendicular to the vertical axis.
Thus, an Atlanta World structure can be regarded as a composition of multiple Manhattan World structures varying by a rotation around a common (vertical) coordinate axis.
Besides in the context of indoor reconstruction, the Manhattan World assumption, as well as the weaker Atlanta World assumption have been used in a range of other application fields such as point cloud segmentation \citep{straub_et_al_2014, kim_et_al_2017, straub_et_al_2018}, the extraction of road structures from low-scale airborne images \citep{faber_foerstner_2000} or for stabilization and drift reduction in the context of Visual Odomentry (VO) \citep{saurer_et_al_2012, straub_et_al_2015, straub_et_al_2018} and Simulatenous Localization and Mapping (SLAM) \citep{peasley_et_al_2012,yazdanpour_et_al_2019,li_et_al_2020c,liu_meng_2020}.
 
The fact that a given indoor reconstruction approach relies on the Manhattan World assumption does not only imply that the building structure to be reconstructed itself must be compliant to the Manhattan World assumption.
Rather, this also implies that the geometric representation of the respective building in the indoor mapping data must be correctly aligned with the coordinate axes in accordance with the definition of the Manhattan World assumption, i.e. that the surfaces pertaining to the three main directions (or six when considering oriented directions) are aligned with the three axes of the coordinate system.

In the context of indoor mapping however, the pose of the captured building structure with respect to the coordinate system does not necessarily fulfill this requirement.
Frequently, the coordinate system is determined by the initial pose of the indoor mapping system at the beginning of the mapping process.
Thus, the orientation of the indoor mapping data can deviate from the Manhattan World assumption by a rotation around the vertical coordinate axis even if the mapped building structure itself is totally compliant with the Manhattan World assumption.
Moreover, the orientation of the vertical axis itself can also deviate from its optimal orientation according to the Manhattan World assumption, i.e. being perpendicular to horizontal ceiling and floor surfaces.
This is generally not the case when a leveled mounting of the respective indoor mapping sensor is used, e.g. in the case of tripod-mounted systems like TLS or trolley-based systems.
In the case of hand-held or head-worn indoor mapping systems where a perfectly leveled orientation at the start of the indoor mapping process cannot be guaranteed, an eventual misalignment of the indoor mapping data with respect to the vertical coordinate axis needs to be taken into account.

Aligning an indoor mapping dataset with the coordinate axes - horizontally and depending on the used indoor mapping system also vertically - is thus a necessary preprocessing step for automated indoor reconstruction approaches that rely on the Manhattan World assumption.
Moreover, a suchlike alignment process - also known as pose normalization - can still be a reasonable choice, even if the respective indoor reconstruction method does not presuppose a Manhattan World compliant building structure.
This is for instance the case, when a respective indoor reconstruction approach makes use of a voxel grid or octree representation of the input data \citep{fichtner_et_al_2017,gorte_et_al_2019,coudron_et_al_2020,huebner_et_al_2020b}.
Even if a voxel-based indoor reconstruction approach is able to handle building structures deviating from the Manhattan World assumption, having room surfaces aligned with the coordinate axes and thus with the voxel grid will result in a cleaner and visually more appealing reconstruction in voxel space.
Furthermore, spatially discretizing data which is not aligned with the coordinate axes can lead to aliasing effects that can impede a successful reconstruction process \citep{martens_blankenbach_2019,martens_blankenbach_2020,xu_et_al_2021}.
Besides, pose normalization often - but not necessarily always, depending on the respective building structure - results in a minimal axis-aligned bounding box circumscribing the indoor mapping data and thus to reduced memory size of the voxel grid structure.

Lastly, pose normalization of indoor mapping data can also be of benefit in the context of the co-registration of multiple datasets representing the same indoor environment that are to be aligned with each other \citep{wijmans_furukawa_2017,chen_et_al_2020,huang_et_al_2020}.
The respective datasets to be aligned can be acquired by different sensor systems or at different times, e.g. in the context of change detection \citep{bassier_et_al_2019c,koeva_et_al_2019,maalek_et_al_2019}.
While pose normalization with respect to a Manhattan World structure does not entirely solve this problem as an ambiguity of rotations of multiples of \SI{90}{\degree} around the vertical axis remains, it nonetheless can be reasonable to apply pose normalization when co-registering indoor mapping datasets as it reduces the problem to finding the correct of only four possible states per dataset. 

The same arguments speaking in favor of pose normalization - even if an indoor reconstruction approach does not necessarily depend on it - also hold for the case of building structures that are only partly compliant to the Manhattan World assumption.
Thus, a pose normalization approach should be robust against a substantial amount of the given indoor mapping geometries deviating from the Manhattan World structure of the building.
Particularly in the case of building environments that contain multiple Manhattan World structures (i.e. Atlanta World), the dominant Manhattan World structure (e.g. in terms of the largest fraction of supporting geometries) should be used for alignment with the coordinate axes.
In situations, where multiple Manhattan World structures have about the same support, it might be reasonable to detect them all and create multiple solutions for a valid pose normalization.

In a more general context, a range of pose normalization approaches have been presented that aim at aligning arbitrary three-dimensional objects with the coordinate axes. 
These objects do not necessarily represent building structures \citep{kazhdan_2007, papadakis_et_al_2007, chaouch_verroust-blondet_2008, fu_et_al_2008, lian_et_al_2008, chaouch_verroust-blondet_2009, lian_et_al_2010, sfikas_et_al_2011}.
These approaches are mainly motivated by the need to design rotation invariant shape descriptors in the context of shape retrievable, i.e. finding all similar three-dimensional objects to a given query shape from a large database of 3D objects \citep{zhang_lu_2001, tangelder_veltkamp_2007}.

In this context, variations of the Principal Component Analysis (PCA) algorithm \citep{jolliffe_cadima_2016} are often made use of \citep{kazhdan_2007,papadakis_et_al_2007,chaouch_verroust-blondet_2008,chaouch_verroust-blondet_2009}.
Also, symmetries in the geometry of the respective object are often exploited as well \citep{chaouch_verroust-blondet_2008, fu_et_al_2008,chaouch_verroust-blondet_2009}.
Other approaches rely on the geometric property of rectilinearity \citep{lian_et_al_2008,lian_et_al_2010} or aim to minimize the size of a surface-oriented bounding box circumscribing the target object \citep{sfikas_et_al_2011}.

More specifically concerning building structures, a recent pose normalization approach makes use of Point Density Histograms, discretizing and aggregating the points of an indoor mapping point cloud along the direction of one of the horizontal coordinate axes \citep{martens_blankenbach_2019,martens_blankenbach_2020}.
The optimal horizontal alignment of the point cloud is determined by maximizing the size and distinctness of peaks in this histogram varying with the rotation around the vertical axis.

Other approaches, including the one proposed in this paper, do not discretize the data with respect to their position but with respect to their orientation \citep{okorn_et_al_2010, khoshelham_diaz_vilarino_2014, diaz-vilarino_et_al_2015, czerniawski_et_al_2016}.
This is conducted on the Extended Gaussian Image \citep{horn_et_al_1984} which consists of the normal vectors of the individual indoor mapping geometries projected on the unit sphere.
Besides its application in the context of pose normalization, the Extended Gaussian Image is also frequently applied to the segmentation of point clouds \citep{wang_et_al_2013c, straub_et_al_2014, shui_et_al_2016, straub_et_al_2018, zhao_et_al_2020} or plane detection \citep{limberger_oliveira_2015}, particularly with regard to building structures.

In a straight-forward approach for instance, the points in the Extended Gaussian Image are subjected to a k-Means clustering \citep{macqueen_1967,lloyd_1982} to determine three clusters corresponding to the main directions of the Manhattan World structure while disregarding the absolute orientation of the normal vectors (i.e. projecting them all in the same hemisphere) \citep{khoshelham_diaz_vilarino_2014, diaz-vilarino_et_al_2015}.
This however is not robust to deviations of the indoor mapping point cloud from an ideal Manhattan World structure.
In contrast, using DBSCAN \citep{ester_et_al_1996} for clustering on the Extended Gaussian Image has been proposed \citep{czerniawski_et_al_2016} which is more robust as it does not fix the number of clusters to exactly three.
This allows for the presence of surfaces deviating from an ideal Manhattan World system.
The proposed approach however only aims at detecting dominant planes to remove them from the point cloud and does not assemble the detected orientation clusters to Manhattan World structures.
In another approach, dominant horizontal directions are detected by projecting the normal vectors to the horizontal plane and binning the resulting angles to a horizontal reference coordinate axis in a similar manner to the approach presented in this paper \citep{okorn_et_al_2010}.

All of the approaches mentioned above only concern themselves with determining an orientation around the vertical axis to achieve an alignment of the Manhattan World structure of an indoor mapping dataset with the coordinate axes.
To the best of our knowledge, no approach on pose normalization of indoor mapping point clouds or triangle meshes has yet been proposed that aims at determining an optimal alignment with respect to the orientation of the vertical axis as well.
Furthermore, the presented approaches do not address the topic of robustness to deviations of the respective building structure from an ideal Manhattan World scenario or the presence of multiple Manhattan World structures in the same building.

In this work, we present a novel pose normalization method for indoor mapping point clouds and triangle meshes that is robust to the represented building structures being only partly compliant to the Manhattan World assumption.
In case there are multiple major Manhattan World structures present in the data, the dominant one is detected and used for alignment.
Besides the horizontal alignment of the Manhattan World structure with the coordinate system axes, vertical alignment is also supported for cases where the deployed indoor mapping system is not leveled and the resulting dataset is thus misaligned with respect to the vertical coordinate axis.
In this context, we presume that the indoor mapping dataset is coarsely leveled to within $\pm$\SI{30}{\degree} of the optimal vertical direction which can usually be expected to be the case for hand-carried or head-worn mobile indoor mapping systems.
We furthermore presuppose the individual indoor mapping geometries to have normal vectors which however do not need to be consistently oriented and can thus be easily determined as a pre-processing step for point clouds while triangle meshes do already have normal vectors inherent in the geometries of the individual triangles.
Our implementation of the proposed pose normalization approach along with the code for the presented quantitative evaluation on publicly available indoor mapping datasets will be made available to the public upon acceptance for publication.

The presented approach for pose normalization is described in \autoref{sec:method} along with a method to resolve the ambiguity of a rotation of multiples of \SI{90}{\degree} around the vertical axis and the procedure applied for quantitative evaluation.
The results of this evaluation procedure applied to several publicly available indoor mapping point clouds and triangle meshes are subsequently presented in \autoref{sec:results} and discussed in further detail in \autoref{sec:discussion}.
Finally, we close in \autoref{sec:conclusions} with concluding remarks and an outlook on future research.
    \section{Materials and Methods}
\label{sec:method}
    
    In this section, we present a novel method for automatic pose normalization of indoor mapping point clouds or triangle meshes which represent building structures that are at least partially compliant to the Manhattan World assumption.
    The presented method aims at rotating the given indoor mapping geometries to a pose with respect to the surrounding coordinate system for which the largest possible fraction of normal vectors is aligned with the three Cartesian coordinate axes.
    This comprises an optional leveling step to orient horizontal surfaces like floors and ceilings to be orthogonal to a chosen vertical axis if this is not already achieved by the data acquisition process (e.g. by using leveled tripod or trolley mounted acquisition systems).
    Subsequently, a second step determines the optimal rotation angle around this vertical axis in order to align the largest possible fraction of the building surfaces with the horizontal pair of orthogonal coordinate axes.
    
    The presented method is applicable to all kind of indoor mapping point clouds and triangle meshes.
    However, we assume the individual geometric primitives comprising the input data to have normal vectors.
    While these are intrinsically given for the individual triangles comprising a triangle mesh, the individual points of indoor mapping point clouds do not generally have normal vectors.
    These can however be easily determined by means of established methods like \citep{mitra_nguyen_2003,boulch_marlet_2012,yu_et_al_2019,sanchez_et_al_2020} which we assume in this work as a necessary preprocessing step.
    Note that these normal vectors need not be oriented, i.e. only their direction is of importance.
    Furthermore, we assume the input data to be at least coarsely levelled, i.e. we assume the represented building structures to be coarsely aligned with the vertical axis within the range of $\pm$\SI{30}{\degree}.
    
    In the following, $\Vec{n}_{i}$ denotes the $i$-th normal vector of $N$ given input geometries (i.e. points or triangles) while $\langle\cdot,\cdot\rangle$ denotes the dot product of two 3D vectors.
    Furthermore, the vector determining the vertical axis is denoted by $\Vec{z}$.
    However, it needs to be stated that this vector need not necessarily equal \(\begin{smallmatrix}(0 & 0 & 1)\end{smallmatrix}^{T}\).
    It can be chosen freely in accordance with the intended coordinate system.
    However, it must coincide within $\pm$\SI{30}{\degree} with the current vertical orientation of the input data.
    Similarly, a horizontal axis $\Vec{x}$ orthogonal to the configured $\Vec{z}$-axis is to be chosen.
    Lastly, the second horizontal axis completing the Cartesian coordinate system must not be explicitly stated but can be determined as 
    
    \begin{equation}	
        \Vec{y}=\Vec{z}\times\Vec{x}
    \end{equation}
    Again, note that the horizontal axes need not necessarily equal \(\begin{smallmatrix}(1 & 0 & 0)\end{smallmatrix}^{T}\) and \(\begin{smallmatrix}(0 & 1 & 0)\end{smallmatrix}^{T}\).
    
    In the following, \autoref{sec:method_rotationAroundVerticalAxis} first presents the proposed method for determining an optimal rotation around the vertical axis in order to horizontally align the indoor mapping data with the coordinate system in case the dataset is already vertically aligned in relation to the vertical axis.
    A suitable method for ensuring this vertical alignment that can be applied as a preprocessing step to datasets that are only coaresly aligned with the vertical direction ($\pm$\SI{30}{\degree}) is subsequently presented in \autoref{sec:method_orientationOfVerticalAxis}.
    As the proposed method for determining the rotation around the vertical axis is ambiguous with regard to multiples of \SI{90}{\degree}, \autoref{sec:method_unambiguosness} presents an approach to solve this ambiguity.
    Lastly, \autoref{sec:method_evaluation} presents the evaluation methodology applied in this study.
    
    \subsection{Rotation around the vertical axis}
    \label{sec:method_rotationAroundVerticalAxis}
    
        In this section, we preliminarily assume that the given indoor mapping data (comprised of triangles or points) is already leveled with regard to a chosen vertical axis $\Vec{z}$ (that does not necessarily need to equal \(\begin{smallmatrix}(0 & 0 & 1)\end{smallmatrix}^{T}\)).
        Thus, only one rotation angle around this vertical axis is to be determined in order to align the two horizontal axes of the coordinate system with the horizontal directions of the dominant Manhattan World structure underlying the respective building represented by the input data.
        
        In case the given input data is not entirely compliant to the Manhattan World assumption, a best-possible solution in terms of the alignment of all normal vectors with the horizontal coordinate axes is to be found.
        Even indoor mapping data that represents building structures entirely compliant to the Manhattan World assumption can have a significant amount of normal vector directions deviating from the directions of the respective Manhattan World system.
        These deviating normal vector directions can be caused by actual unevenness of walls, by noise inherent in data acquisition and normal determination as well as by clutter like furniture objects being present in the indoor mapping data additionally to the building structure itself.
        
\begin{figure}
	\centering
	\begin{tikzpicture}
        \node at (0,-1.1){         
            \includegraphics[scale=0.35]{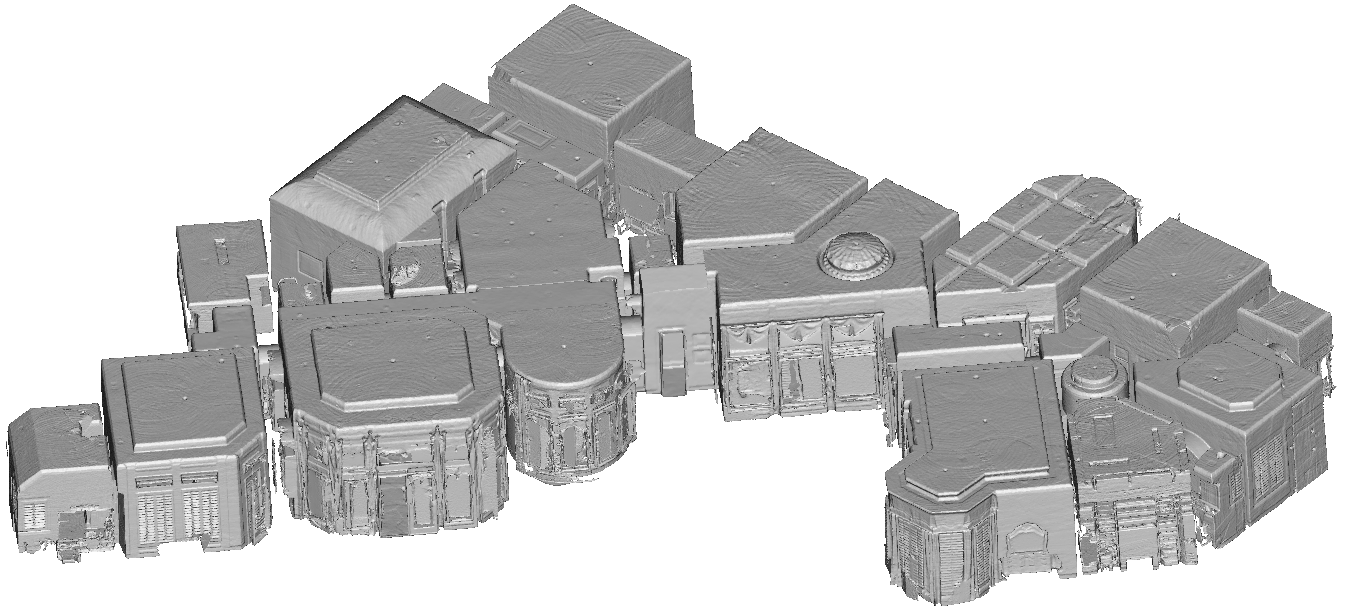}
        };
        \node at (6.5,0){         
            \includegraphics[scale=0.4]{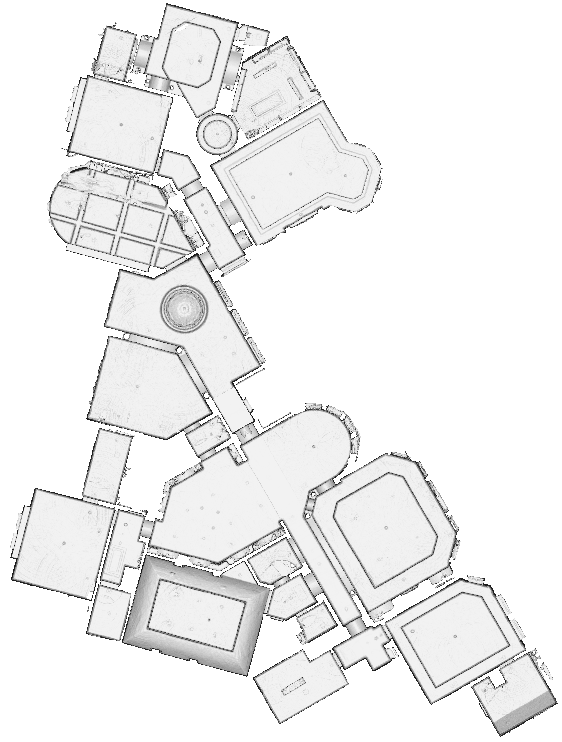}
        };
        \draw [draw=red, line width=1pt] (4.5,-2.65) rectangle (8.5,2.65);
        \begin{scope}[shift={(0.4,-3.5)}]
            \draw [draw=green, line width=1pt, rotate=30] (4.6,-3.1) rectangle (8.42,2.7);
        \end{scope}
    \end{tikzpicture}
    \caption { 
        Exemplary triangle mesh of a building with multiple Manhattan World systems (dataset 'mJXqzFtmKg4' from Matterport3D \citep{chang_et_al_2017}). The green bounding box on the top-down-view on the right-hand side illustrates the alignment along the dominant Manhattan World structure considered as ground truth pose while the red bounding box illustrates the pose rotated by \SI{30}{\degree} around the vertical axis as exemplarily used in \autoref{sec:method_rotationAroundVerticalAxis}.
    }
  	\label{fig:method_rotationAroundVerticalAxis_exampleBuilding}
\end{figure}

        Besides being robust against these restrictions, the presented method is also applicable to building structures that are only partially Manhattan World conform.
        Building structures with multiple Manhattan World systems like the one depicted in \autoref{fig:method_rotationAroundVerticalAxis_exampleBuilding} are aligned according to the respective Manhattan World system supported by the largest fraction of normal vector directions.
        
\begin{figure}
	\centering
	\subfigure {
      \includegraphics[width=6cm]{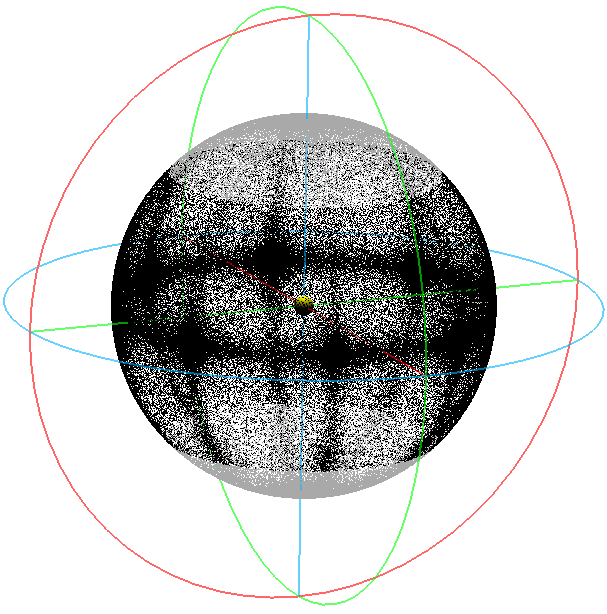}
    }
    \subfigure {
      \includegraphics[width=6cm]{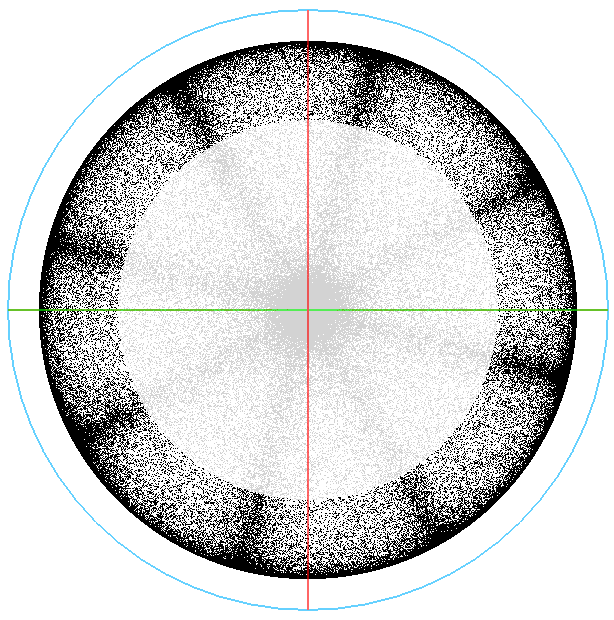}
    }
    \caption {
        The normal vectors $\Vec{n}_{i}$ of the triangle mesh shown in \autoref{fig:method_rotationAroundVerticalAxis_exampleBuilding} visualized as Extended Gaussian Image (thinned out by factor 25 for the sake of visibility). The normal vectors $\Vec{n}^{h}_{i}$ that are horizontal within the range of $\pm$\SI{45}{\degree} are visualized in black while the others are visualized in gray. The coordinate axes are visualized in red for $\Vec{x}$, green for $\Vec{y}$ and blue for the vertical axis $\Vec{z}$.
    }
  	\label{fig:method_rotationAroundVerticalAxis_normals}
\end{figure}

        Thus, the task at hand is to determine an angle of rotation around the vertical axis $\Vec{z}$ that leads to the largest positive fraction of normal vectors being aligned with the horizontal axes $\Vec{x}$ and $\Vec{y}$.
        To this aim, we first filter the normal vectors that can be considered coarsely horizontal with respect to the vertical axis $\Vec{z}$.
        For this, we consider all $N^{h}$ normal vectors $\Vec{n}^{h}_{i}$ that are within the range of $\pm$\SI{45}{\degree} of a horizontal orientation, thus
        
        \begin{equation}	
            \SI{45}{\degree}\leqslant|\sphericalangle(\Vec{n}_{i},\Vec{z})|\leqslant\SI{135}{\degree}
        \end{equation}
        where $\sphericalangle(\cdot,\cdot)$ denotes the smallest angle between two 3D vectors with respect to any rotation axis.
        For the indoor mapping mesh depicted in \autoref{fig:method_rotationAroundVerticalAxis_exampleBuilding}, the corresponding horizontal normal vectors $\Vec{n}^{h}_{i}$ are depicted in the form of an Extended Gaussian Image in \autoref{fig:method_rotationAroundVerticalAxis_normals}.
        In this example, the triangle mesh of \autoref{fig:method_rotationAroundVerticalAxis_exampleBuilding} is rotated by \SI{30}{\degree} around the vertical axis relative to the ground truth pose aligned to the dominant Manhattan World structure.
        
        These horizontal normal vectors $\Vec{n}^{h}_{i}$ can subsequently be projected in the horizontal plane formed by the horizontal axes $\Vec{x}$ and $\Vec{y}$ by
        
        \begin{equation}	
            \Vec{\tilde n}^{h}_{i}=\Vec{n}^{h}_{i}-\langle\Vec{n}^{h}_{i},\Vec{z}\rangle\Vec{z}
        \end{equation}
        where their respective angles to the reference direction of $\Vec{x}$ around $\Vec{z}$ as axis of rotation
        
        \begin{equation}	
            \gamma _{i}=\sphericalangle_{\Vec{z}}(\Vec{\tilde n}^{h}_{i},\Vec{x})=\arctan{\frac{\langle\Vec{z},\Vec{\tilde n}^{h}_{i}\times\Vec{x}\rangle}{\langle\Vec{\tilde n}^{h}_{i},\Vec{x}\rangle}}\in[\SI{-180}{\degree},\SI{180}{\degree})
        \end{equation}
        can be determined.
        
        The problem at hand can be formulated as determining the rotation angle $\gamma\in[\SI{0}{\degree},\SI{90}{\degree})$ around the vertical axis that minimizes the sum of angular distances of each horizontal normal vector to the respectively nearest horizontal coordinate axis, i.e.:
            \begin{equation}	
                \label{eq:method_rotationAroundVerticalAxis_argmin}
                \gamma=\argmin_{\hat\gamma\in[\SI{0}{\degree},\SI{90}{\degree})} \sum^{N^{h}}_{i=0} w_{i}\min
                \begin{Bmatrix}
                     |\hat\gamma - \gamma_{i}| \\
                     |\hat\gamma - \gamma_{i} + \SI{90}{\degree}| \\
                     |\hat\gamma - \gamma_{i} + \SI{180}{\degree}| \\
                     |\hat\gamma - \gamma_{i} - \SI{90}{\degree}| \\
                \end{Bmatrix}
            \end{equation}
            
        Here, the angular distances of each angle $\gamma_{i}$ to the nearest horizontal axis are weighted by factor $w_{i}$.
        This factor can be constantly set to 1 for the points of an indoor mapping point cloud.
        In the case of triangle meshes however, it allows to weigh the individual triangles by their respective area as larger triangles imply a larger quantity of points in a corresponding point cloud representation.
        
        \autoref{eq:method_rotationAroundVerticalAxis_argmin} is not analytically solvable.
        It can however be solved numerically by derivative-free minimization methods like e.g. Brent Minimization \citep{brent_1973}.
        This, however, does not scale well with the size of the input data, as all the angles derived from the horizontal normal vectors need to be iterated in each step of the respective numeric method.
        And - particularly in the case of indoor mapping point clouds - the amount of geometric primitives and thus of angles to be processed can reach a tremendous size.
        
        Thus, in this work, we propose an approach that discretizes the input data into a one-dimensional grid of fixed resolution by means of which the angle of rotation for aligning the input data with the horizontal coordinate system can be determined non-iteratively in one step. In this context, a resolution of \SI{1}{\degree} proved to be suited for a coarse initial determination of the rotation angle for horizontal alignment that can subsequently be refined.
        For each angle $\gamma_{i}$, the respective grid cell is determined which is incremented by the respective weight $w_{i}$, which again is constantly 1 for points of point clouds but in the case of triangle meshes weights the respective angle by the area of the corresponding triangle.
        
\begin{figure}
	\centering
    \begin{tikzpicture}
        \node at (0,0){         
            \includegraphics[width=5cm]{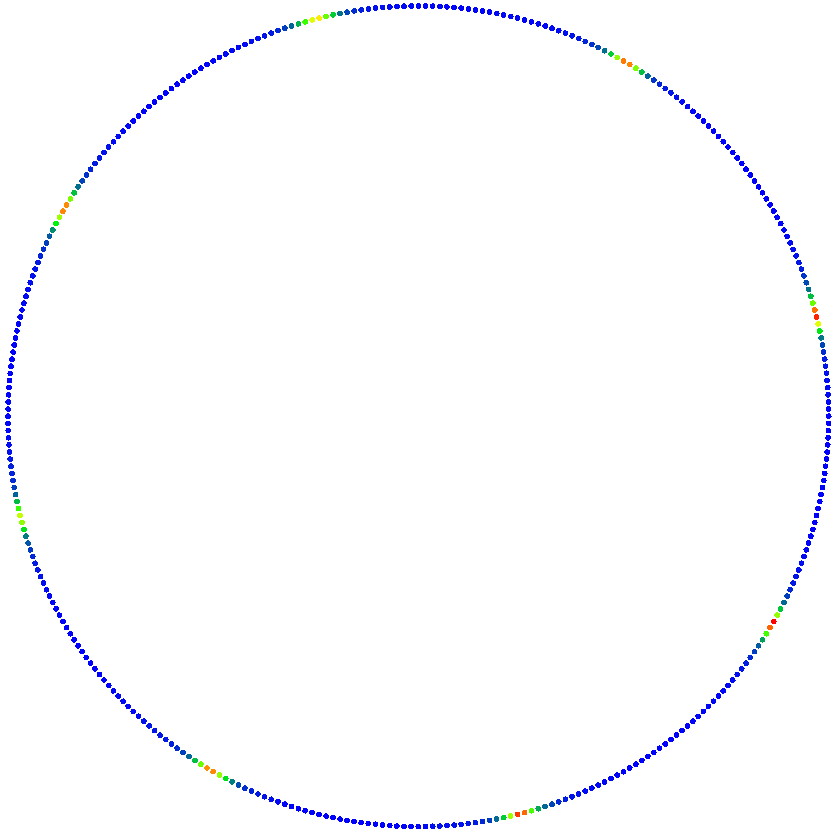}
        };
        \draw [densely dotted] (-2.7, 0) -- (2.7, 0);
        \draw [densely dotted] (0, -2.7) -- (0, 2.7);
        \draw [->, line width=1.5pt, red] (0, 0) -- (1.5, 0);
        \draw [->, line width=1.5pt, black!30!green] (0, 0) -- (0, 1.5);
        \filldraw [blue] (0, 0) circle (0.05);
        \node [red] at (1.5, -0.25) {$\Vec{x}$};
        \node [black!30!green] at (0.25, 1.5) {$\Vec{y}$};
        \node at (2.2, 0) {\SI{0}{\degree}};
        \node at (0, -2.2) {\SI{-90}{\degree}};
        \node at (0, 2.2) {\SI{90}{\degree}};
        \node at (-1.75, 0.2) {\SI{180}{\degree}/};
        \node at (-1.9, -0.2) {\SI{-180}{\degree}};
        \begin{scope}[shift={(3.5,-2.5)}]
            \begin{axis}[
                    xtick={-180,-90,0,90,180},
                    ytick={0,10,20,30},
                    xlabel near ticks,
                    ylabel near ticks,
                    xlabel={$\gamma_{i}$ {[}\textdegree{]}},
                    ylabel={$\sum w_{i}$},
                    ylabel style={
                        rotate=-90,
                        xshift=0.65cm,
                        yshift=2.4cm
                    }
                ]
                \addplot+ table [
                        x index=0, 
                        y index=1, 
                        col sep=semicolon
                    ] {figures/data/grid_horizontal_full.csv};
            \end{axis}
        \end{scope}
    \end{tikzpicture}
    \caption {
        Visualization of a one-dimensional \SI{360}{\degree} grid corresponding to \autoref{fig:method_rotationAroundVerticalAxis_exampleBuilding}. The grid cells contain the summarized weights $w_{i}$ of the contained angles $\gamma_{i}$ with value colorization ranging from blue for low values over green and yellow to red for large values.
    }
  	\label{fig:method_rotationAroundVerticalAxis_grid_full}
\end{figure}
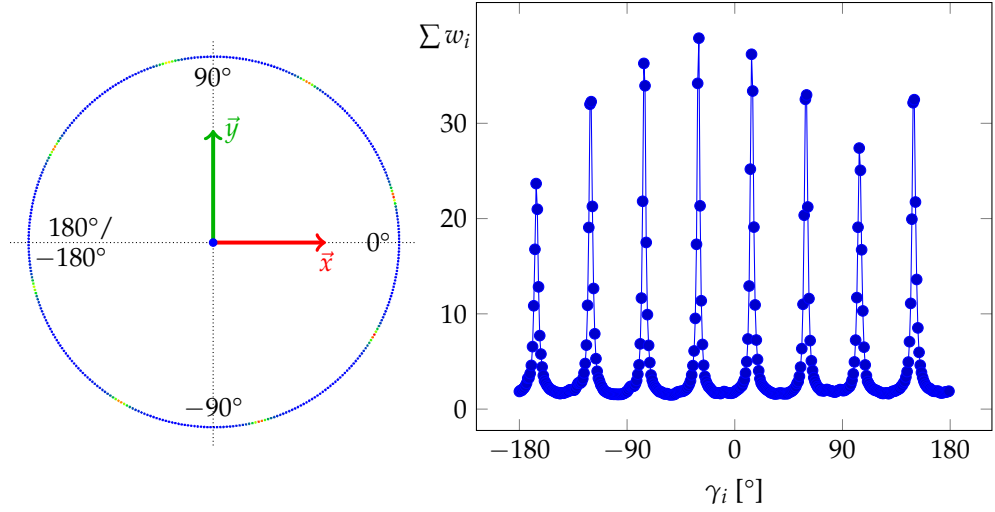

        \autoref{fig:method_rotationAroundVerticalAxis_grid_full} visualizes a suchlike one-dimensional grid representation of the horizontal angles $\gamma_{i}$ over the full circle of \SI{360}{\degree} for the mesh presented in \autoref{fig:method_rotationAroundVerticalAxis_exampleBuilding}.
        The peaks in the summarized weights per grid cell correspond to the eight horizontal main directions of the two Manhattan World systems present in the dataset depicted in \autoref{fig:method_rotationAroundVerticalAxis_exampleBuilding}.
        
        To decide about the dominant of the two Manhattan World systems involved and to determine the corresponding rotation angle for an alignment of the input data with it, the weights of the involved grid cells need to be summarized over all peaks pertaining to the same Manhattan World system.
        To this end, the peaks belonging to the same Manhattan World system and thus having an angular difference of a multiple of \SI{90}{\degree} between each other need to be identified and associated.
        Thus, we map the angles $\gamma_{i}\in[\SI{-180}{\degree},\SI{180}{\degree})$ to $[\SI{0}{\degree},\SI{90}{\degree})$ by
        
        \begin{equation}
            \gamma^{\ast}_{i}=
            \begin{cases}
                \gamma_{i}+\SI{180}{\degree} & \gamma_{i} < \SI{0}{\degree} \\
                \gamma_{i} & else \\
            \end{cases}
            \in[\SI{0}{\degree},\SI{180}{\degree})
        \end{equation}
        and 
        
        \begin{equation}
            \Tilde{\gamma}_{i}=
            \begin{cases}
                \gamma^{\ast}_{i}-\SI{90}{\degree} & \gamma^{\ast}_{i} > \SI{90}{\degree} \\
                \gamma^{\ast}_{i} & else \\
            \end{cases}
            \in[\SI{0}{\degree},\SI{90}{\degree})
        \end{equation}
        
\begin{figure}
	\centering
	\begin{tikzpicture}
        \node at (0,0){         
            \includegraphics[width=5cm]{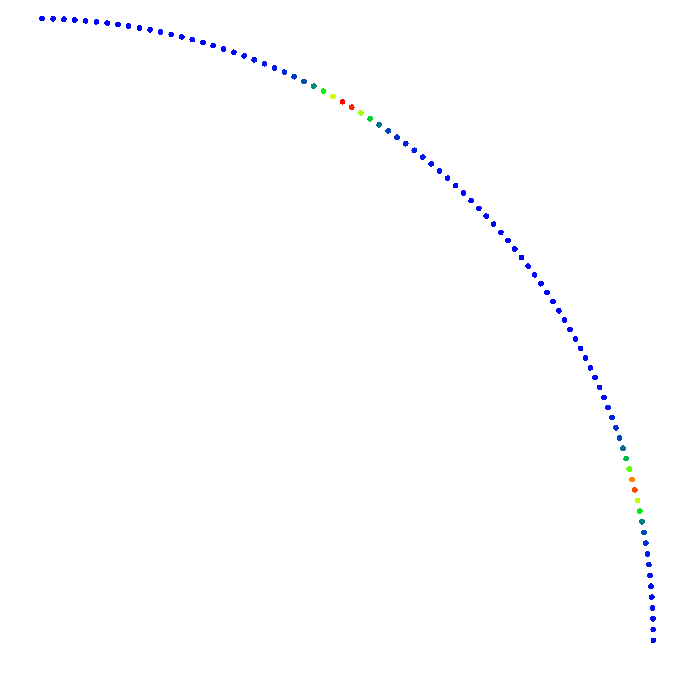}
        };
        \draw [densely dotted] (-2.21, -2.26) -- (2.7, -2.26);
        \draw [densely dotted] (-2.21, -2.26) -- (-2.21, 2.7);
        \draw [->, line width=1.5pt, red] (-2.21, -2.26) -- (-0.71, -2.26);
        \draw [->, line width=1.5pt, black!30!green] (-2.21, -2.26) -- (-2.21, -0.76);
        \filldraw [blue] (-2.21, -2.26) circle (0.05);
        \node [red] at (-0.71, -2) {$\Vec{x}$};
        \node [black!30!green] at (-1.9, -0.76) {$\Vec{y}$};
        \node at (2.1, -2) {\SI{0}{\degree}};
        \node at (-1.88, 2) {\SI{90}{\degree}};
        \draw [gray, densely dotted] (3.5, 1.45) -- (10.38, 1.45);
        \node [gray] at (9.5, 1.62) {\footnotesize{Threshold}};
        \node [gray] at (9.3, 1.27) {\footnotesize{(75\% of max.)}};
        \begin{scope}[shift={(3.5,-2.5)}]
            \begin{axis}[
                    xtick={0,15,30,45,60,75,90},
                    xlabel near ticks,
                    ylabel near ticks,
                    xlabel={$\Tilde{\gamma}_{i}$ {[}\textdegree{]}},
                    ylabel={$\sum w_{i}$},
                    ylabel style={
                        rotate=-90,
                        xshift=0.8cm,
                        yshift=2.4cm
                    }
                ]
                \addplot+ table [
                        x index=0, 
                        y index=1, 
                        col sep=semicolon
                    ] {figures/data/grid_horizontal.csv};
            \end{axis}
        \end{scope}
    \end{tikzpicture}
    \caption {
        Visualization of a one-dimensional \SI{90}{\degree} grid corresponding to \autoref{fig:method_rotationAroundVerticalAxis_exampleBuilding}. The grid cells contain the summarized weights $w_{i}$ of the contained angles $\Tilde{\gamma}_{i}$ with value colorization ranging from blue for low values over green and yellow to red for large values.
    }
  	\label{fig:method_rotationAroundVerticalAxis_grid}
\end{figure}
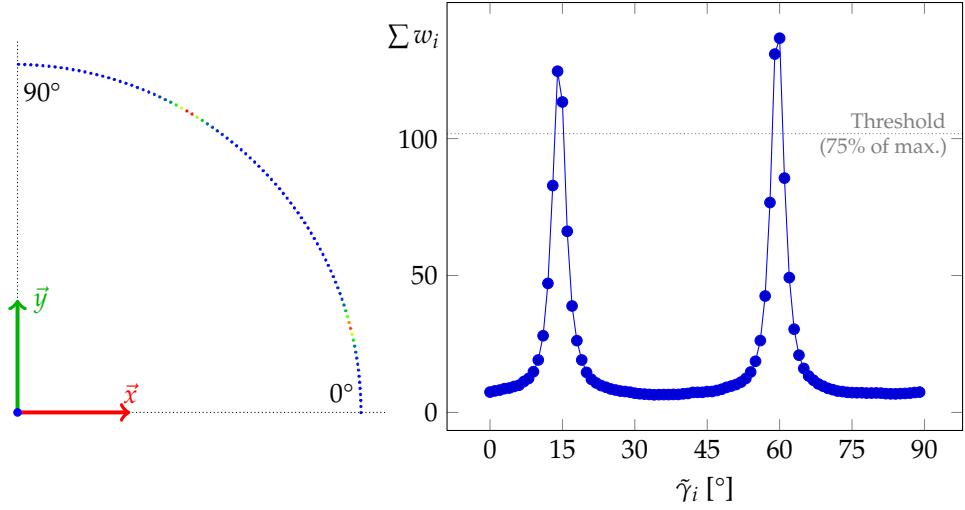

        The discretized grid representation of the angles $\Tilde{\gamma}_{i}\in[\SI{0}{\degree},\SI{90}{\degree})$ thus needs only a quarter of the size in comparison to discretizing the angles $\gamma_{i}\in[\SI{-180}{\degree},\SI{180}{\degree})$ with the same resolution.
        Furthermore, the resulting grid as visualized in \autoref{fig:method_rotationAroundVerticalAxis_grid} enables the coarse initial determination of the rotation angle $\gamma$.
        To this end, the weight sums per grid cell are thresholded with a threshold value of 0.75 times the maximal weight sum of the whole grid and subsequently clustered.
        While clustering, the fact that clusters can extend over the discontinuity between \SI{0}{\degree} and \SI{90}{\degree} needs to be taken account of.
        
\begin{figure}
	\centering
	\subfigure[
	        Faces corresponding to the largest peak at \SI{60}{\degree} in \autoref{fig:method_rotationAroundVerticalAxis_grid} determining the dominanting Manhattan World structure.]{
        \label{fig:method_rotationAroundVerticalAxis_horizontalFaces_cluster1}
        \includegraphics[width=14cm]{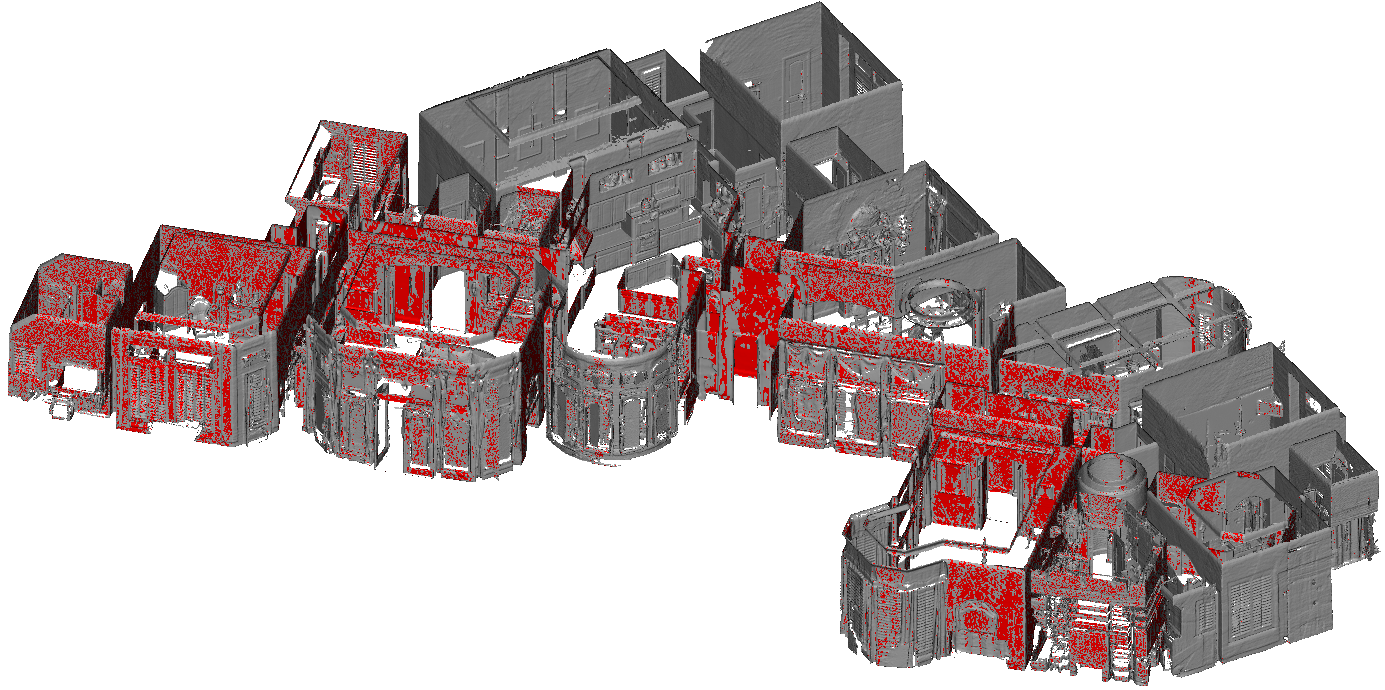}
    }
    \subfigure[
	        Faces corresponding to the minor peak at \SI{15}{\degree} in \autoref{fig:method_rotationAroundVerticalAxis_grid}.]{
        \label{fig:method_rotationAroundVerticalAxis_horizontalFaces_cluster2}
        \includegraphics[width=14cm]{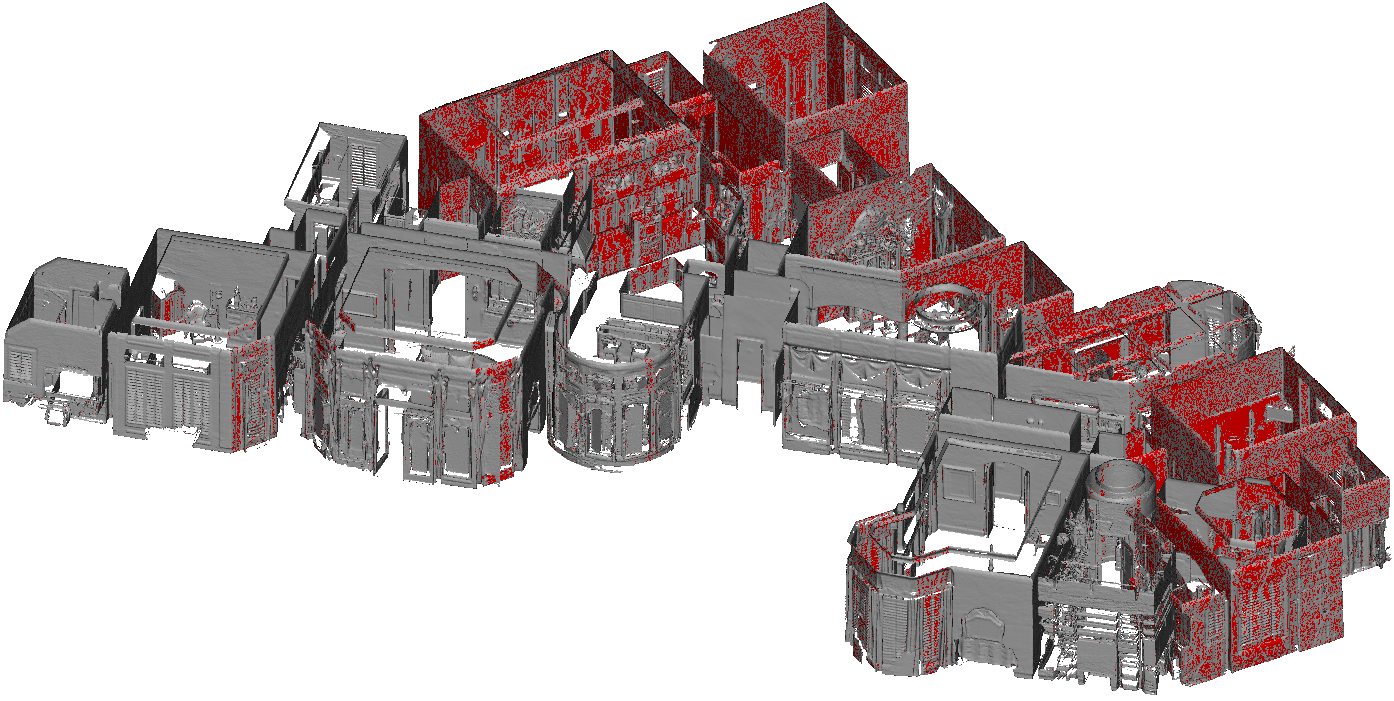}
    }
    \caption {
        The horizontal faces of the triangle mesh presented in \autoref{fig:method_rotationAroundVerticalAxis_exampleBuilding} corresponding to the horizontal normal vectors $\Vec{n}^{h}_{i}$.
        The faces corresponding to the two peaks shown in \autoref{fig:method_rotationAroundVerticalAxis_grid} are depicted in red respectively.
    }
  	\label{fig:method_rotationAroundVerticalAxis_horizontalFaces}
\end{figure}

        Finally, the grid cell cluster with the largest weight summarized over the contained cells is selected and $\gamma$ is determined as the weighted average of the angle values corresponding to the cluster cells (with \SI{1}{\degree} resolution) weighted by their respective weight sum values.
        \autoref{fig:method_rotationAroundVerticalAxis_horizontalFaces_cluster1} shows the horizontal triangle mesh faces of \autoref{fig:method_rotationAroundVerticalAxis_exampleBuilding} corresponding to the largest peak at \SI{60}{\degree} in \autoref{fig:method_rotationAroundVerticalAxis_grid} that determines the dominant Manhattan World system of that dataset.
        The faces corresponding to the second peak at \SI{15}{\degree} in  \autoref{fig:method_rotationAroundVerticalAxis_grid} are visualized in \autoref{fig:method_rotationAroundVerticalAxis_horizontalFaces_cluster2}.
        
        The resulting value for $\gamma$ can subsequently be further refined by determining the weighted median over all $\Tilde{\gamma}_{i}$ within a certain angular distance of the initial value for $\gamma$ while applying the weights $w_{i}$.
        A threshold of \SI{5}{\degree} was found to be suitable for this task.
        
        Finally, the indoor mapping data can be rotated by the thus refined angle $\gamma$ around the vertical axis to achieve the alignment of the building geometry with the horizontal coordinate axes.
        In the case of a triangle mesh, it is sufficient to rotate the vertices of the triangles as the respective normal vectors of the rotated triangles can be calculated on the basis of the triangle geometry.
        In the case of point clouds however, the respective normal vectors of the points need to be explicitly updated along with the coordinates of the points.
    
    \subsection{Orientation of the vertical axis}
    \label{sec:method_orientationOfVerticalAxis}
        
        In the preceding \autoref{sec:method_rotationAroundVerticalAxis}, the rotation around the vertical axis was determined under the assumption that the vertical axis is perfectly leveled with respect to the building structure, i.e. that it is orthogonal to horizontal floor and ceiling surfaces.
        In the case of tripod mounted indoor mapping systems like terrestrial laser scanners, this assumption is justified as these devices are typically leveled before usage.
        However, in the case of mobile indoor mapping systems like hand-carried or head-worn devices, this is generally not the case.
        In these cases, the coordinate system of the indoor mapping data is often defined by the initial pose of the mobile mapping device when starting the data acquisition process.
        In consideration of typical usage postures of such mobile systems, it can be assumed that the respective vertical axis of the coordinate system is still roughly pointing upwards $\pm$\SI{30}{\degree}.
        If this is not the case, a coarse leveling within this range can easily be conducted manually.
        
        To justify the assumption made in the previous section, this section presents an approach for automatically leveling indoor mapping point clouds or triangle meshes where a chosen vertical axis $\Vec{z}$ corresponds coarsely within $\pm$\SI{30}{\degree} with the actual upwards direction of the building structure standing orthogonally on horizontal floor surfaces.
        As in the preceding section, the input data for conducting this alignment of the input mapping data with the coordinate system are again the $N$ normal vectors $\Vec{n}_{i}$ of the individual geometric primitives comprising the indoor mapping data (i.e. points or triangles). 
        
        Analogous to \autoref{eq:method_rotationAroundVerticalAxis_argmin}, we can formulate the task of vertically aligning the indoor mapping geometries with the coordinate system axis $\Vec{z}$ as
        
        \begin{equation}
            \label{eq:method_orientationOfVerticalAxis_argmin}
            \begin{pmatrix}
                \alpha \\ 
                \beta
            \end{pmatrix}=\argmin_{\hat\alpha,\hat\beta\in[\SI{-30}{\degree},\SI{30}{\degree}]}\sum^{N^{v}}_{i=0} w_{i}\min\begin{Bmatrix}
                 |\sphericalangle(R(\hat\alpha,\hat\beta)\Vec{n}^{v}_{i},\Vec{z})| \\
                 |\sphericalangle(R(\hat\alpha,\hat\beta)\Vec{n}^{v}_{i},\Vec{z}) - \SI{180}{\degree}|
            \end{Bmatrix}
        \end{equation}
        where $\Vec{n}^{v}_{i}$ are the $N^{v}$ normal vectors that are vertically oriented within the range
        
        \begin{equation}
            |\sphericalangle(\Vec{n}_{i},\Vec{z})|\leqslant\SI{40}{\degree}\wedge|\sphericalangle(\Vec{n}_{i},\Vec{z})|\geqslant\SI{140}{\degree}
        \end{equation}
        and $w_{i}$ again is a weighting factor being constant for points of a point cloud but corresponding to the respective triangle area for the faces of a triangle mesh.
        Furthermore, $R(\alpha,\beta)$ denotes a $3\times3$ rotation matrix determined by two rotation angles $\alpha$ and $\beta$ around the two horizontal coordinate axes $\Vec{x}$ and $\Vec{y}$ respectively.
        
\begin{figure}
	\centering
	\begin{tikzpicture}
        \node at (0,-0.2){         
            \includegraphics[scale=0.42]{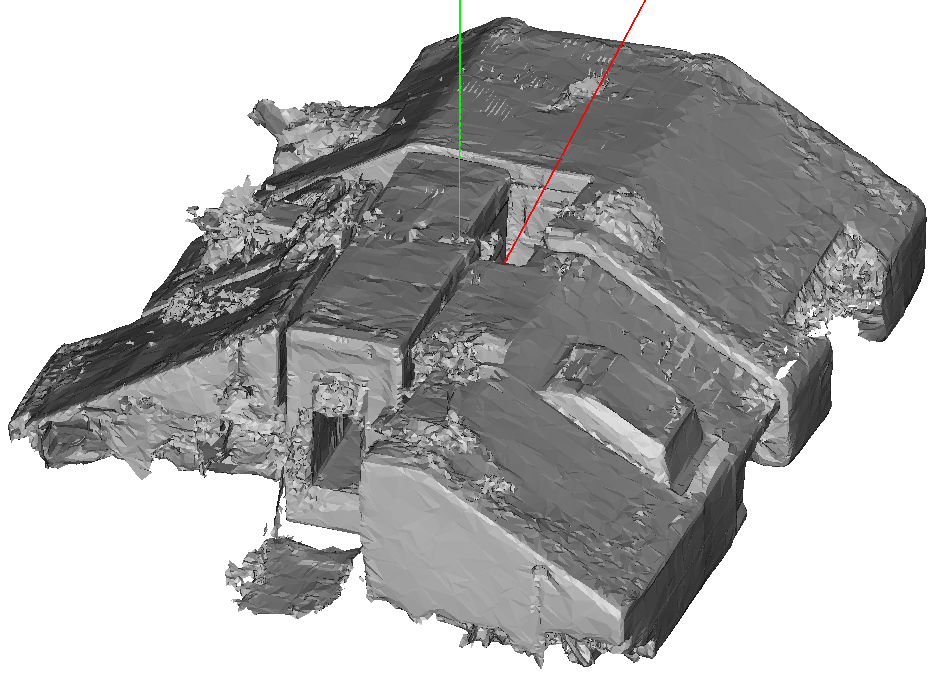}
        };
        \node at (6.8,2){         
            \includegraphics[scale=0.32]{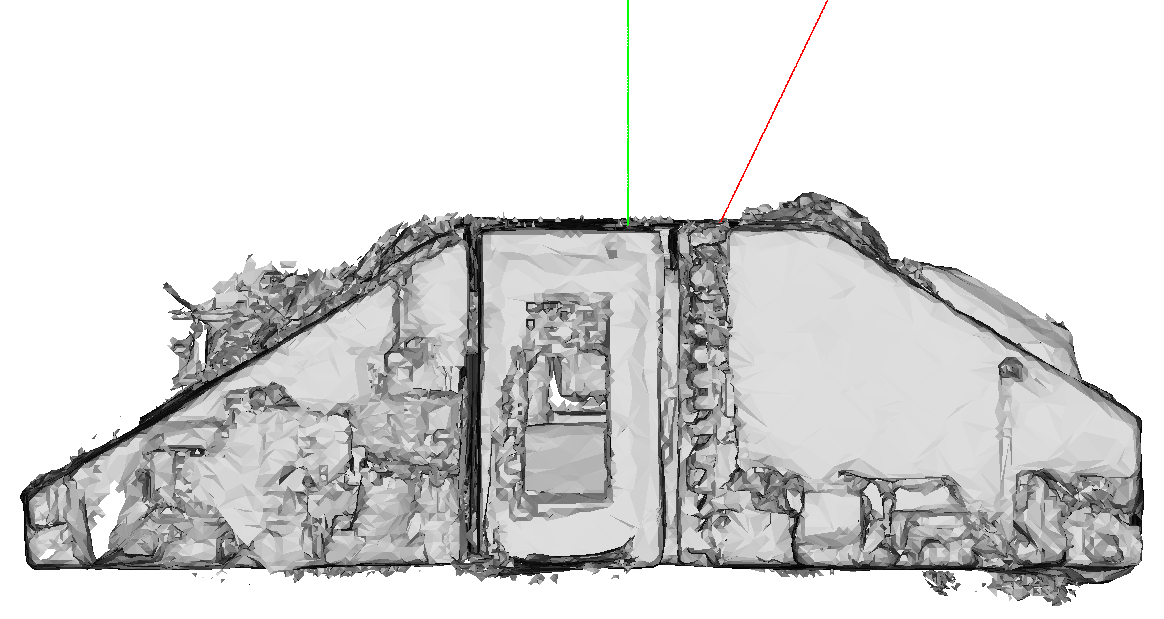}
        };
        \node at (6.8,-1.5){         
            \includegraphics[scale=0.32]{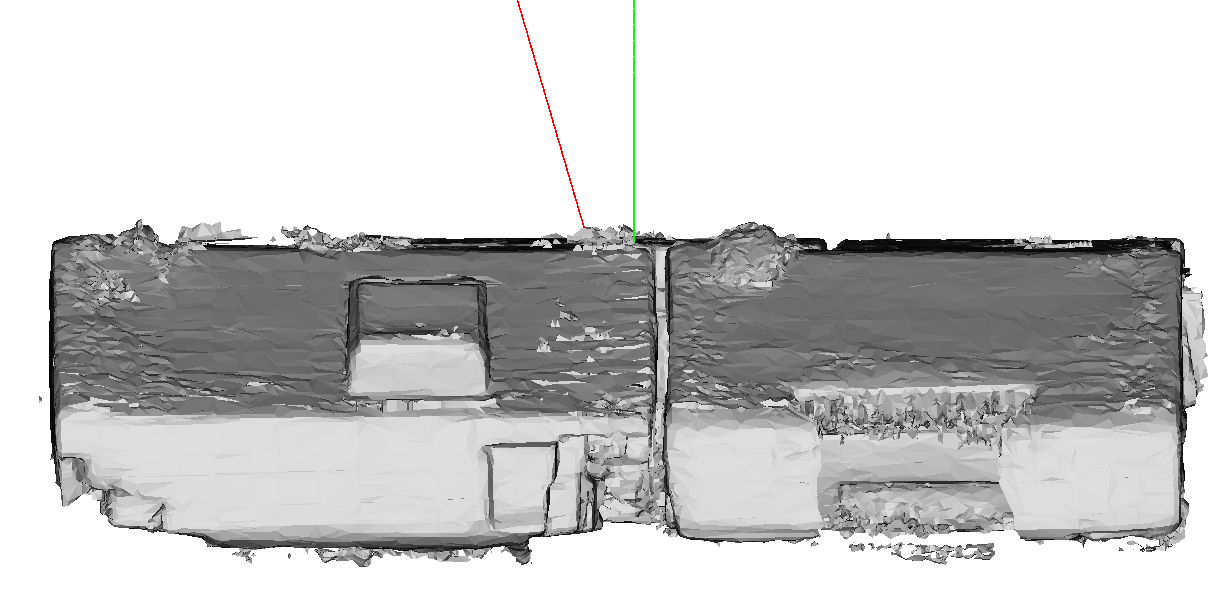}
        };
    \end{tikzpicture}
    \caption { 
        Exemplary triangle mesh of a building with partially slanted ceiling (dataset 'Attic' from \citep{huebner_et_al_2021}). The green line visualizes the reference orientation of the vertical axis considered as ground truth while the red line visualizes the vertical axis rotated \SI{-25}{\degree} around the horizontal $\Vec{x}$ axis and \SI{15}{\degree} around the horizontal $\Vec{y}$ axis as examplarily used in \autoref{sec:method_orientationOfVerticalAxis}.
    }
  	\label{fig:method_orientationOfVerticalAxis_exampleBuilding}
\end{figure}

        Thus, the aim of \autoref{eq:method_orientationOfVerticalAxis_argmin} is two find the optimal vertical axis $\Vec{z}^{\ast}$ as a vector
        
        \begin{equation}
            \Vec{z}^{\ast}=R(\alpha, \beta)\Vec{z}
        \end{equation}
        in the initially given coordinate system that has a minimal sum of angles to the vertical normals $\Vec{n}^{v}_{i}$.
        This optimal vertical axis $\Vec{z}^{\ast}$ as well as the initial vertical axis $\Vec{z}$ are exemplarily depicted in \autoref{fig:method_orientationOfVerticalAxis_exampleBuilding} for a building with slanted ceilings only coarsely aligned with the actual vertical direction.
        
        As it already was the case with \autoref{eq:method_rotationAroundVerticalAxis_argmin} in \autoref{sec:method_rotationAroundVerticalAxis}, \autoref{eq:method_orientationOfVerticalAxis_argmin} is not analytically solvable and solving it numerically is all the more inefficient as this time, a two-dimensional minimization is concerned.
        Thus, as in the case of determining the rotation angle around the vertical axis in \autoref{sec:method_rotationAroundVerticalAxis}, we again seek to formulate the problem at hand as the task of searching a maximum peak within a discrete grid representation of the relevant input elements.
        
        The relevant input elements in this case are the three-dimensional vertical normal vectors $\Vec{n}^{v}_{i}$.
        However, the problem at hand is actually two-dimensional as a rotation around the two horizontal axes $\Vec{x}$ and $\Vec{y}$ by the rotation angles $\alpha$ and $\beta$ is sufficient for aligning the vertical axis $\Vec{z}$ with the optimal vertical direction $\Vec{z}^{\ast}$.
        
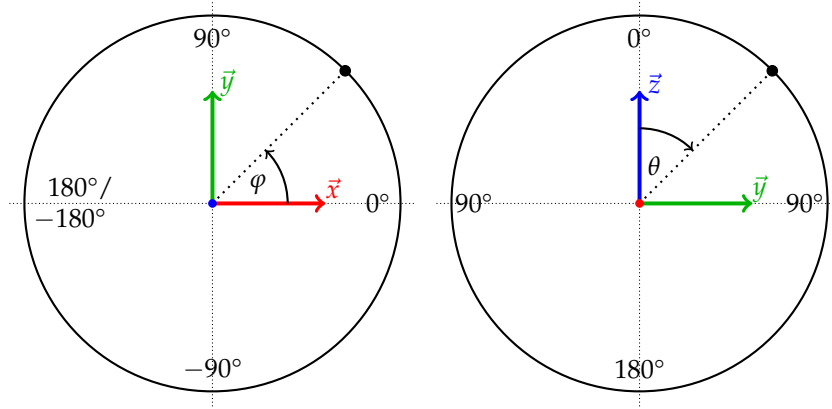
\begin{figure}
	\centering
	\subfigure {
	    \begin{tikzpicture}
    	    \draw [densely dotted] (-2.7, 0) -- (2.7, 0);
            \draw [densely dotted] (0, -2.7) -- (0, 2.7);
            \draw[thick] (0, 0) circle (2.5);
            \draw [->, line width=1.5pt, red] (0, 0) -- (1.5, 0);
            \draw [->, line width=1.5pt, black!30!green] (0, 0) -- (0, 1.5);
            \node [red] at (1.6, 0.2) {$\Vec{x}$};
            \node [black!30!green] at (0.2, 1.6) {$\Vec{y}$};
            \node at (2.2, 0) {\SI{0}{\degree}};
            \node at (0, -2.2) {\SI{-90}{\degree}};
            \node at (0, 2.2) {\SI{90}{\degree}};
            \node at (-1.75, 0.2) {\SI{180}{\degree}/};
            \node at (-1.9, -0.2) {\SI{-180}{\degree}};
            \draw[->, thick] (1,0) arc (0:45:1);
            \node at (0.6, 0.25) {$\varphi$};
            \draw [thick, dotted] (0, 0) -- (1.765, 1.765);
            \filldraw [blue] (0, 0) circle (0.05);
            \filldraw (1.765, 1.765) circle (0.07);
	    \end{tikzpicture}
    }
    \subfigure {
        \begin{tikzpicture}
    	    \draw [densely dotted] (-2.7, 0) -- (2.7, 0);
            \draw [densely dotted] (0, -2.7) -- (0, 2.7);
	        \draw[thick] (0, 0) circle (2.5);
	        \draw [->, line width=1.5pt, black!30!green] (0, 0) -- (1.5, 0);
            \draw [->, line width=1.5pt, blue] (0, 0) -- (0, 1.5);
	        \node [black!30!green] at (1.6, 0.2) {$\Vec{y}$};
	        \node [blue] at (0.2, 1.6) {$\Vec{z}$};
	        \node at (2.2, 0) {\SI{90}{\degree}};
            \node at (0, -2.2) {\SI{180}{\degree}};
            \node at (0, 2.2) {\SI{0}{\degree}};
            \node at (-2.2, 0) {\SI{90}{\degree}};
            \draw[->, thick] (0,1) arc (90:45:1);
            \node at (0.2, 0.5) {$\theta$};
            \draw [thick, dotted] (0, 0) -- (1.765, 1.765);
            \filldraw [red] (0, 0) circle (0.05);
            \filldraw (1.765, 1.765) circle (0.07);
	    \end{tikzpicture}
    }
    \caption {
        Azimuth $\varphi$ and inclination $\theta$.
    }
  	\label{fig:method_orientationOfVerticalAxis_sphericalCoordinates}
\end{figure}

        In an alternative formulation, this can also be considered as the task of finding the position of the optimal vertical direction $\Vec{z}^{\ast}$ on the surface of a unit sphere, i.e. within the Extended Gaussian Image.
        The orientation of a normal vector with respect to the coordinate system can be expressed via the polar angles azimuth
        
        \begin{equation}
            \varphi_{i}=\arctan{\frac{\langle\Vec{n}_{i},\Vec{y}\rangle}{\langle\Vec{n}_{i},\Vec{x}\rangle}}\in[\SI{-180}{\degree},\SI{-180}{\degree})
        \end{equation}
        and inclination
        
        \begin{equation}
            \theta_{i}=\arccos{\langle\Vec{n}_{i},\Vec{z}\rangle}\in[\SI{0}{\degree},\SI{-180}{\degree})
        \end{equation}
        indicating the position of a respective normal vector $\Vec{n}_{i}$ on the unit sphere.
        The definition of azimuth and inclination with respect to the coordinate system is further illustrated in \autoref{fig:method_orientationOfVerticalAxis_sphericalCoordinates}.
        
\begin{figure}
	\centering
	\subfigure {
	    \begin{tikzpicture}
    	    \node at (0,-0.2){         
                 \includegraphics[scale=0.42]{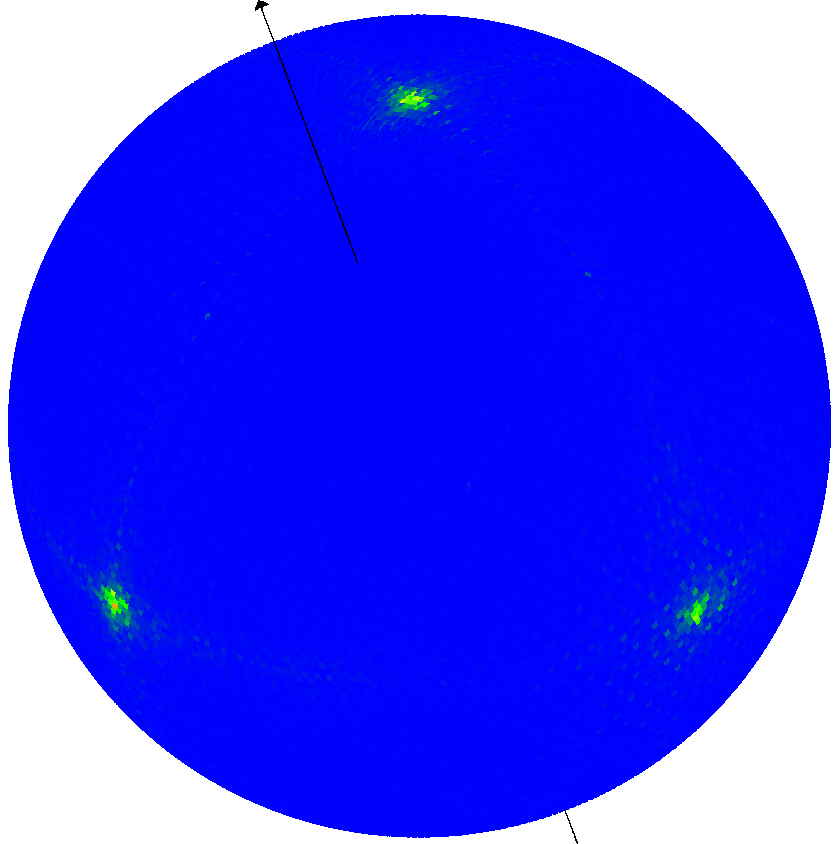}
            };
            \node at (-1.4, 2.8) {$\Vec{z}$};
            \node at (0.4, 1.7) {\textbf{Floor}};
            \node at (-1.6, -1.5) {\textbf{Wall\textsubscript{1}}};
            \node at (1.3, -1.5) {\textbf{Wall\textsubscript{2}}};
	    \end{tikzpicture}
    }
    \subfigure {
        \begin{tikzpicture}
    	    \node at (0,-0.2){         
                 \includegraphics[scale=0.42]{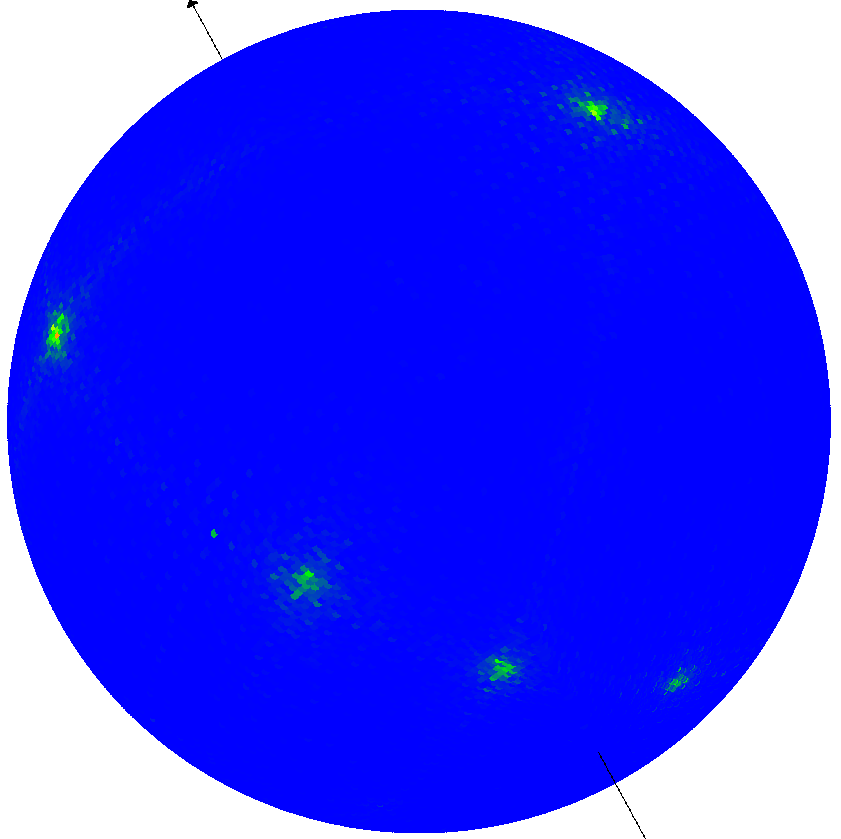}
            };
            \node at (-1.4, 2.8) {$\Vec{z}$};
            \node at (-2, 0.5) {\textbf{Wall\textsubscript{1}}};
            \node at (0.8, 1.8) {\textbf{Wall\textsubscript{2}}};
            \node [align=left] at (0.7, -1.4) {\footnotesize{\textbf{Horizontal}}\\\footnotesize{\textbf{ Ceiling}}};
            \node [align=right] at (-1.7, -1.4) {\footnotesize{\textbf{Slanted}}\\\footnotesize{\textbf{ Ceiling\textsubscript{1}}}};
            \node [align=right] at (1.95, -1.6) {\footnotesize{\textbf{Slanted}}\\\footnotesize{\textbf{ Ceiling\textsubscript{2}}}};
	    \end{tikzpicture}
    }
    \caption {
        Azimuth/inclination grid of \SI{1}{\degree} resolution over the whole surface of the unit sphere corresponding to \autoref{fig:method_orientationOfVerticalAxis_exampleBuilding}.
        The grid cells contain the summarized weights $w_{i}$ of the contained normal vectors $\Vec{n}_{i}$ at polar angles $(\varphi_{i}, \theta_{i})$ with value colorization ranging from blue for low values over green and yellow to red for large values.
    }
  	\label{fig:method_orientationOfVerticalAxis_grid_full}
\end{figure}
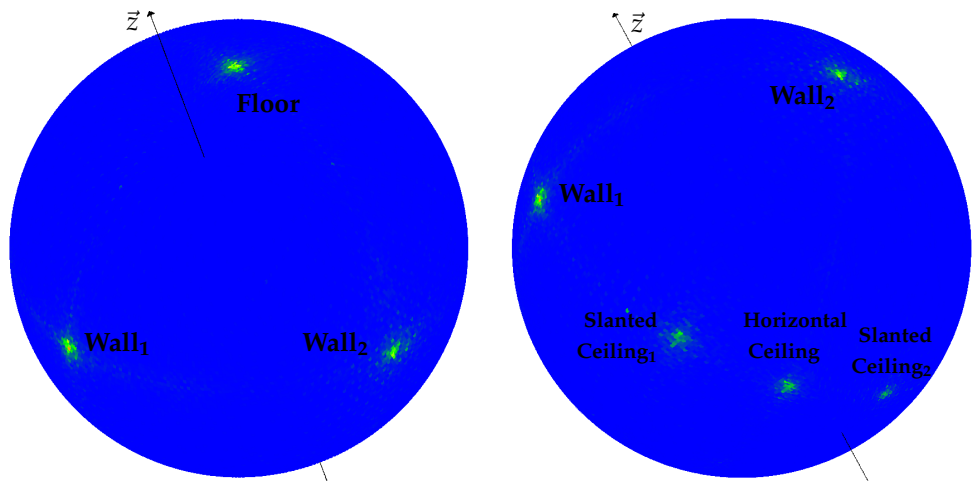

        This representation allows us to construct a two-dimensional azimuth/inclination grid analogous to the approach presented in \autoref{sec:method_rotationAroundVerticalAxis} whose cells are weighted by the summarized weights $w_{i}$ of the contained normal vectors $\Vec{n}_{i}$.
        A suchlike grid of a resolution of \SI{1}{\degree} extending over the whole unit sphere is depicted in \autoref{fig:method_orientationOfVerticalAxis_grid_full} corresponding to the exemplary case presented in \autoref{fig:method_orientationOfVerticalAxis_exampleBuilding}.
        
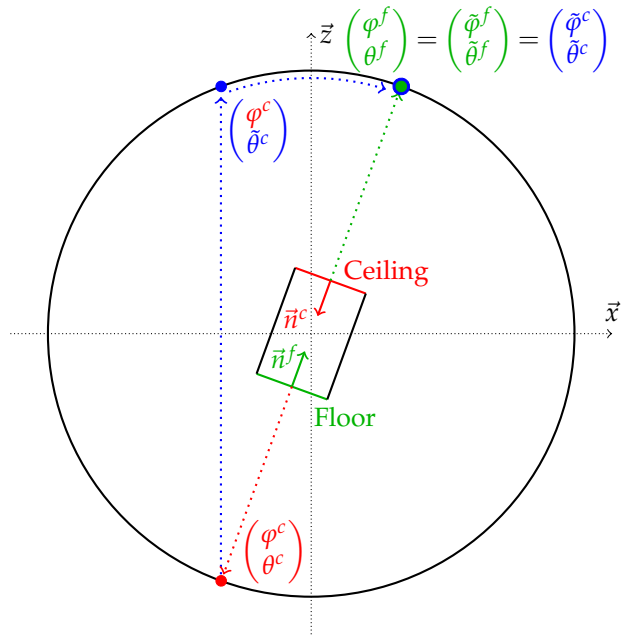
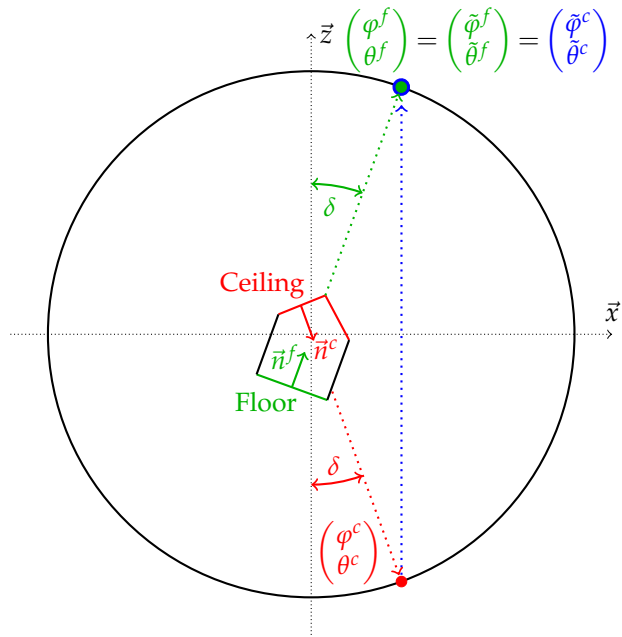
\begin{figure}
	\centering
	\subfigure[Generally, points corresponding to opposing normal vectors get transformed to the same point.] {
	    \label{fig:method_orientationOfVerticalAxis_sphericalCoordinates_transformed_a}
	    \begin{tikzpicture}
    	    \draw[->, densely dotted] (-4, 0) -- (4, 0);
            \draw[->, densely dotted] (0, -4) -- (0, 4);
            \draw[thick] (0, 0) circle (3.5);
            \node at (4, 0.3) {$\Vec{x}$};
            \node at (0.2, 4) {$\Vec{z}$};
            \draw[->, dotted, thick, blue] (-1.2, -3.3) -- (-1.2, 3.15);
            \begin{scope}[rotate around={-20:(0,0)}]
                \draw[thick, red] (-0.5,0.75) -- (0.5,0.75);
                \draw[thick, black!30!green] (-0.5,-0.75) -- (0.5,-0.75);
                \draw[thick] (-0.5,-0.75) -- (-0.5,0.75);
                \draw[thick] (0.5,-0.75) -- (0.5,0.75);
                \draw[->, thick, red] (0,0.75) -- (0,0.25);
                \draw[->, thick, black!30!green] (0,-0.75) -- (0,-0.25);
                \draw[->, dotted, thick, red] (0,-0.75) -- (0,-3.41);
                \draw[->, dotted, thick, black!30!green] (0,0.75) -- (0,3.41);
                \filldraw[red] (0,-3.5) circle (0.07);
                \filldraw[blue] (0,3.5) circle (0.11);
                \filldraw[black!30!green] (0,3.5) circle (0.07);
            \end{scope}
            \begin{scope}[rotate around={20:(0,0)}]
                \filldraw[blue] (0,3.5) circle (0.07);
            \end{scope}
            \draw[<-, dotted, thick, blue] ([shift=(73:3.4)]0,0) arc (73:109:3.4);
            \node[red] at (1, 0.8) {Ceiling};
            \node[black!30!green] at (0.45, -1.1) {Floor};
            \node[red] at (-0.2, 0.2) {$\Vec{n}^{c}$};
            \node[black!30!green] at (-0.35, -0.3) {$\Vec{n}^{f}$};
            \node[red] at (-0.5, -2.9) {$\begin{pmatrix}\varphi^{c}\\\theta^{c}\end{pmatrix}$};
            \node at (-0.7, 2.7) {$\color{blue}\begin{pmatrix}\textcolor{red}{\varphi^{c}} \\ \Tilde{\theta}^{c}\end{pmatrix}$};
            \node at (2.2, 3.9) {$\color{black!30!green}\begin{pmatrix}\varphi^{f} \\ \theta^{f}\end{pmatrix}\textcolor{black}{=}\begin{pmatrix}\Tilde{\varphi}^{f} \\ \Tilde{\theta}^{f}\end{pmatrix}\textcolor{black}{=}\begingroup\color{blue}\begin{pmatrix}\Tilde{\varphi}^{c} \\ \Tilde{\theta}^{c}\end{pmatrix}\endgroup$};
	    \end{tikzpicture}
    }
    \subfigure[In case the vertical axis $\Vec{z}$ is the angle bisector between the directions of two normal vectors (same angle $\delta$ to $\Vec{z}$ axis), these get transformed to the same point even if they are not opposed. This needs to be dealt with by means of a cluster analysis per $(\Tilde{\varphi}, \Tilde{\theta})$ grid cell.] {
        \label{fig:method_orientationOfVerticalAxis_sphericalCoordinates_transformed_b}
        \begin{tikzpicture}
    	    \draw[->, densely dotted] (-4, 0) -- (4, 0);
            \draw[->, densely dotted] (0, -4) -- (0, 4);
	        \draw[thick] (0, 0) circle (3.5);
	        \node at (4, 0.3) {$\Vec{x}$};
            \node at (0.2, 4) {$\Vec{z}$};
            \begin{scope}[rotate around={-20:(0,0)}]
                \draw[thick, red] (-0.5,0.1) -- (0,0.55);
                \draw[thick, red] (0.5,0.1) -- (0,0.55);
                \draw[thick, black!30!green] (-0.5,-0.75) -- (0.5,-0.75);
                \draw[thick] (-0.5,-0.75) -- (-0.5,0.1);
                \draw[thick] (0.5,-0.75) -- (0.5,0.1);
                \filldraw[blue] (0,3.5) circle (0.11);
                \filldraw[black!30!green] (0,3.5) circle (0.07);
                \draw[->, thick, black!30!green] (0,-0.75) -- (0,-0.25);
                \draw[->, dotted, thick, black!30!green] (0,0.55) -- (0,3.41);
            \end{scope}
            \filldraw[red] (1.2,-3.29) circle (0.07);
            \draw[->, dotted, thick, blue] (1.2, -3.2) -- (1.2, 3.05);
            \begin{scope}[rotate around={20:(1.2,-3.29)}]
                \draw[->, thick, red] (1.2,0.62) -- (1.2,0.12);
                \draw[->, dotted, thick, red] (1.2, -0.6) -- (1.2,-3.20);
            \end{scope}
            \begin{scope}[rotate around={70:(0,0)}]
                \draw[<->, thick, black!30!green] (2,0) arc (0:20:2);    
            \end{scope}
            \begin{scope}[rotate around={-70:(0,0)}]
                \draw[<->, thick, red] (2,0) arc (0:-20:2);    
            \end{scope}
            \node[black!30!green] at (0.25, 1.7) {$\delta$};
            \node[red] at (0.3, -1.75) {$\delta$};
            \node[red] at (-0.65, 0.65) {Ceiling};
            \node[black!30!green] at (-0.6, -0.9) {Floor};
            \node[red] at (0.2, -0.2) {$\Vec{n}^{c}$};
            \node[black!30!green] at (-0.35, -0.3) {$\Vec{n}^{f}$};
            \node[red] at (0.5, -2.9){$\begin{pmatrix}\varphi^{c}\\\theta^{c}\end{pmatrix}$};
            \node at (2.2, 3.9) {$\color{black!30!green}\begin{pmatrix}\varphi^{f} \\ \theta^{f}\end{pmatrix}\textcolor{black}{=}\begin{pmatrix}\Tilde{\varphi}^{f} \\ \Tilde{\theta}^{f}\end{pmatrix}\textcolor{black}{=}\begingroup\color{blue}\begin{pmatrix}\Tilde{\varphi}^{c} \\ \Tilde{\theta}^{c}\end{pmatrix}\endgroup$};
	    \end{tikzpicture}
    }
    \caption {
        Transformation of $(\varphi, \theta)$ positions on the whole unit sphere to $(\Tilde{\varphi}, \Tilde{\theta})$ positions on one eighth of the unit sphere by \autoref{eq:method_orientationOfVerticalAxis_transformation_phi} and \autoref{eq:method_orientationOfVerticalAxis_transformation_theta}.
    }
  	\label{fig:method_orientationOfVerticalAxis_sphericalCoordinates_transformed}
\end{figure}

\begin{figure}
	\centering
	\begin{tikzpicture}
	    \node at (0,-0.2){ \includegraphics[scale=0.42]{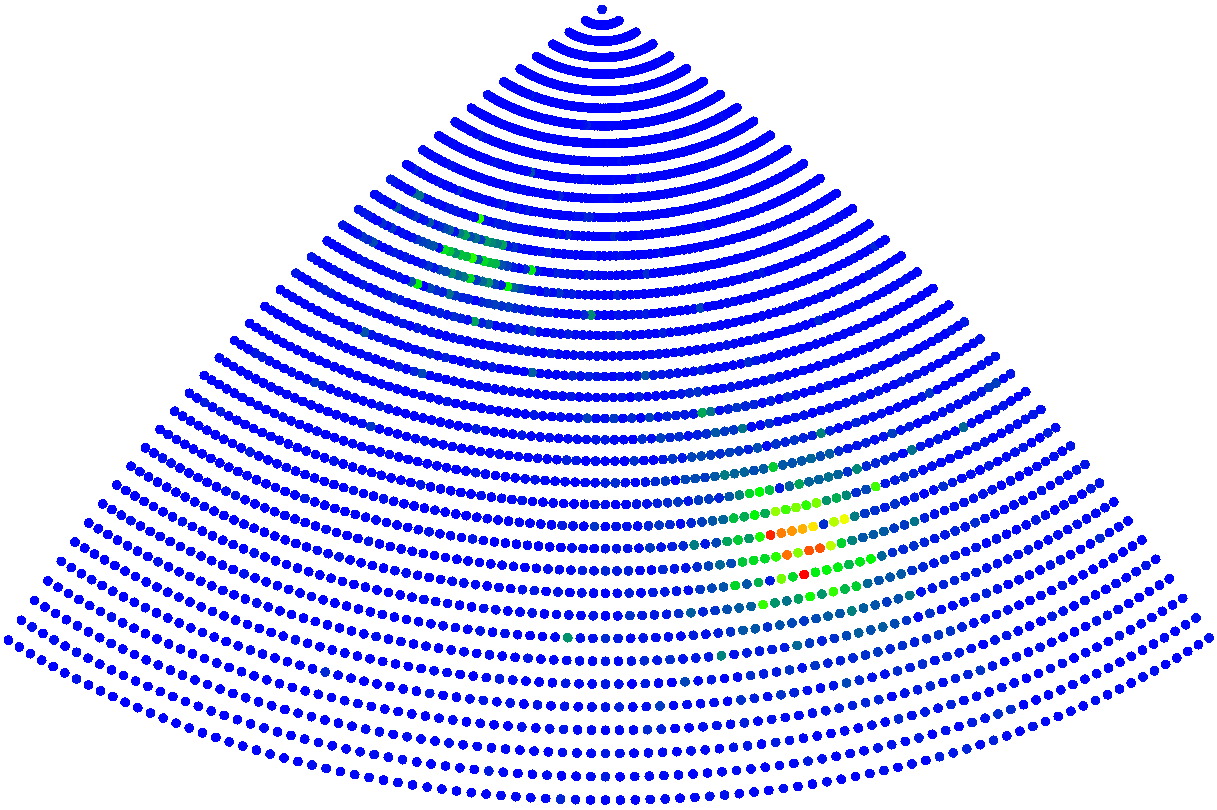}
        };
        \draw[|-|, thick] (-4.6,-2.2) .. controls (-2,-3.7) and (2,-3.7) .. (4.6,-2.2);
        \draw[|-|, thick] (-4.6,-1.8) .. controls (-2.4,1.2) .. (-0.25,2.9);
        \node at (0, -3.6) {$\Tilde{\varphi}$};
        \node at (-2.7, 1) {$\Tilde{\theta}$};
        \node at (-4.6,-2.6) {\SI{0}{\degree}};
        \node at (4.6,-2.6) {\SI{90}{\degree}};
        \node at (-4.7,-1.5) {\SI{40}{\degree}};
        \node at (-0.6,2.9) {\SI{0}{\degree}};
    \end{tikzpicture}
    \caption {
        Transformed azimuth/inclination grid of \SI{1}{\degree} resolution corresponding to \autoref{fig:method_orientationOfVerticalAxis_exampleBuilding}.
        The grid cells contain the summarized weights $w_{i}$ of the contained vertical normal vectors $\Vec{n}^{v}_{i}$ at polar angles $(\Tilde{\varphi}_{i}, \Tilde{\theta}_{i})$ with value colorization ranging from blue for low values over green and yellow to red for large values. The larger peak corresponds to the floor and the horizontal part of the ceiling while the minor peak corresponds to one of the slanted ceiling surfaces.
    }
  	\label{fig:method_orientationOfVerticalAxis_grid}
\end{figure}

        As before in \autoref{sec:method_rotationAroundVerticalAxis}, we want to transform this grid over the full range of the sphere surface to a smaller grid where the weights of cells pertaining to opposing normal vectors get accumulated.
        This is achieved by
        
        \begin{equation}
            \label{eq:method_orientationOfVerticalAxis_transformation_phi}
            \Tilde{\varphi}_{i}=\abs[\Big]{|\varphi_{i}|-\SI{90}{\degree}}\in[\SI{0}{\degree},\SI{90}{\degree})
        \end{equation}
        and
        
        \begin{equation}
            \label{eq:method_orientationOfVerticalAxis_transformation_theta}
            \Tilde{\theta}_{i}=\SI{90}{\degree}-|\theta_{i}-\SI{90}{\degree}|\in[\SI{0}{\degree},\SI{90}{\degree})
        \end{equation}
        while restricting the extension of the grid in the dimension of the inclination to the range of $[\SI{0}{\degree}, \SI{40}{\degree}]$ and thus only considering the vertical normal vectors $\Vec{n}^{v}_{i}$.
        A schematic visualization of this transformation is depicted in \autoref{fig:method_orientationOfVerticalAxis_sphericalCoordinates_transformed_a} while \autoref{fig:method_orientationOfVerticalAxis_grid} shows the resulting two-dimensional azimuth/inclination grid corresponding to the dataset presented in \autoref{fig:method_orientationOfVerticalAxis_exampleBuilding}.
        
        Subsequently, peaks with cell grid weights above a threshold of \SI{75}{\%} of the highest weight value are again clustered like in the case of the one-dimensional grid of \autoref{sec:method_rotationAroundVerticalAxis}.
        While doing so however, not only the azimuth discontinuity between \SI{0}{\degree} and \SI{90}{\degree} needs to be considered, but also the pole point at \SI{0}{\degree} inclination where all azimuth values merge to one and the same grid cell.
        
        While in the case of the one-dimensional grid of \autoref{sec:method_rotationAroundVerticalAxis}, grid cell indices could be directly mapped to angles by multiplication with the grid resolution, here, it is not possible to infer the direction of the optimal vertical axis from grid cell indices as the transformed azimuth values $\Tilde{\varphi}$ are ambiguous by multiples of \SI{90}{\degree}.
        This ambiguousness did also exist in \autoref{sec:method_rotationAroundVerticalAxis}.
        However, it did not affect the correctness of the resulting horizontal alignment as is the case here.
        
        Thus, to be able to deduce correct directions from peaks in the two-dimensional grid, the respective normal vectors $\Vec{n}^{v}_{i}$ need to be hashed per grid cell.
        So, the correct direction of the vertical axis can be initialized by a weighted average of all the hashed normal directions weighted by their respective $w_{i}$ value of the cluster with the largest summarized weight.
        In doing so, normal vectors pointing downwards need to be corrected by inverting the direction to point upwards when calculating the weighted average vector.
        Like in \autoref{sec:method_rotationAroundVerticalAxis}, the initial result is further refined by a weighted median of all normal vectors within $\pm$\SI{5}{\degree} of the coarsely determined resulting vertical axis.
        
        Besides the need to deduce the correct direction from the detected maximum peak grid cells, there is a second reason to hash normal directions per grid cell.
        As illustrated in \autoref{fig:method_orientationOfVerticalAxis_sphericalCoordinates_transformed_b}, two normal vectors that are oriented by the same angle around the vertical axis $\Vec{z}$ in a way that the axis $\Vec{z}$ is the angle bisector between both normals get projected to the same $(\Tilde{\varphi}, \Tilde{\theta})$ grid cell by \autoref{eq:method_orientationOfVerticalAxis_transformation_phi} and \autoref{eq:method_orientationOfVerticalAxis_transformation_theta}.
        On the one hand, this can distort the weight sums of the individual grid cells that are used for peak detection.
        On the other hand, the presence of normal vectors with deviating orientations beyond the ambiguity of $\pm$\SI{180}{\degree} between opposing surfaces can severely distort the initial determination of the vertical direction from the largest peak in the grid.
        
        For this reason, a cluster analysis is conducted among the hashed normal vectors per grid cell.
        In doing so, all the normal vectors in a grid cell are assigned to clusters.
        A normal vector can be assigned to an existing cluster if its direction coincides within $\pm$\SI{2}{\degree} with the average direction of the cluster (with consideration of an ambiguity of $\pm$\SI{180}{\degree}).
        Else, the respective normal vector initializes a new cluster.
        Finally, for each grid cell, only the largest cluster of normals is retained while the others are discarded.
        The grid cell weights and the hashed normal vectors are adapted accordingly.
    
    \subsection{Unambiguousness of the rotation around the vertical axis}
    \label{sec:method_unambiguosness}
        
        The alignment of indoor mapping point clouds or triangle meshes along the coordinate axes as described in the preceding sections \ref{sec:method_rotationAroundVerticalAxis} and \ref{sec:method_orientationOfVerticalAxis} is ambiguous with respect to a rotation around the vertical axes by multiples of \SI{90}{\degree}.
        This is per se not a problem as the aim of the presented approach is to align the indoor mapping data with respect to its Manhattan World structure which inherently implies this ambiguity with respect to four possible rotations around the vertical axis, i.e. all four possible result poses are equally valid with respect to the stated aim.
        
        However, in some situations, it can be desirable to derive an unambiguous pose of the indoor mapping data.
        For instance, this can be the case when multiple indoor mapping results of the same building environment are to be aligned by the proposed method.
        These multiple datasets of the same building can e.g. be obtained by different indoor mapping systems or be acquired at different times in the context of change detection.
        
        For this reason, we present a simple method for resolving the ambiguity in the rotation around the vertical axis by reproducibly choosing one of the four possible horizontal orientations.
        The proposed method presents a straight-forward solution that does not require any semantic interpretation of the indoor mapping data or any elaborate analysis.
        It can however fail in cases of highly symmetric building layouts with respect to its four inherent Manhattan World directions.
        We furthermore presuppose that two datasets to be aligned unambiguously by this method cover approximately the same section of an indoor environment.
        If this is not the case, an approach that incorporates semantic knowledge of the represented indoor environment would be more promising.
        
        Currently however, we propose to resolve the unambiguousness between the four possible horizontal orientations by first aligning the one of the two possible horizontal Manhattan World directions with the chosen reference axis $\Vec{x}$ that corresponds to a larger extent of the bounding box of the respective dataset in this horizontal direction, i.e. the longer horizontal edges of the bounding box should be parallel to the $\Vec{x}$ axis.
        This is quite straight forward but can fail in cases where the bounding box is nearly quadratic.
        
        The ambiguity is now reduced to a rotation of \SI{180}{\degree}.
        To resolve this, we propose to consider the weighted count of indoor mapping geometries in both proximal \SI{10}{\%} sections of the bounding box in $\Vec{x}$ direction and to choose the rotation for which the proximal \SI{10}{\%} section of the bounding box pointing towards the positive $\Vec{x}$ axis has the higher weight sum.
        In this context, the indoor mapping geometries are again weighted by a constant in the case of points of point clouds and by triangle area in the case of triangle mesh faces.
        This approach fails, when the amount of mapped indoor structures in both proximal sections of the bounding box along the $\Vec{x}$ axis is about equal.
    
    \subsection{Evaluation method}
    \label{sec:method_evaluation}
    
        Quantitatively evaluating the proposed method is fortunately quite straight forward as ground truth data can be easily obtained.
        If an indoor mapping dataset is not already correctly aligned with the coordinate system axes in the sense of the aim of this study, it can be aligned manually without great effort.
        A thus aligned dataset can then be rotated to an arbitrary pose within the defined range applicable for the presented method.
        For this a $3\times3$ ground truth rotation matrix $R^{GT}(\alpha,\beta,\gamma)$ is created, determined by the rotation angles $\alpha,\beta\in[\SI{-30}{\degree},\SI{30}{\degree}]$ around the horizontal axes $\Vec{x}$ and $\Vec{y}$ respectively and an arbitrary rotation $\gamma\in[\SI{-180}{\degree},\SI{180}{\degree})$ around the vertical axis $\Vec{z}$.
        For creating $R^{GT}$, the rotation $\gamma$ around the vertical axis is applied first and then successively $\beta$ and $\alpha$ around their respective horizontal axis.
        
        Finally, the method presented in \autoref{sec:method_rotationAroundVerticalAxis} and \autoref{sec:method_orientationOfVerticalAxis} is applied to the rotated dataset which should return the rotated dataset back to its aligned state.
        The resulting $3\times3$ rotation matrix $R^{Test}$ is consituted by
        
        \begin{equation}
            R^{Test}=R^{Test}_{horizontal}R^{Test}_{vertical}
        \end{equation}
        where first $R^{Test}_{vertical}$ is determined by aligning the rotated dataset vertically with the vertical axis as described in \autoref{sec:method_orientationOfVerticalAxis} and then subsequently, the rotation $R^{Test}_{horizontal}$ around the vertical axis is determined as described in \autoref{sec:method_rotationAroundVerticalAxis}.
        
        As an evaluation metric, the angular difference $\delta_{v}$ between the vector of the ground truth axis $\Vec{z}$ and the resulting vector 
        
        \begin{equation}
            \Vec{z}^{Test}=R^{Test}R^{GT}\Vec{z}
        \end{equation}
        is determined by
        
        \begin{equation}
            \delta_{v}=|\sphericalangle(\Vec{z}^{Test},\Vec{z})|
        \end{equation}
        as well as the analogous angular difference $\delta_{h}$ for the horizontal axis $\Vec{x}$.
        In case of the horizontal deviation $\delta_{h}$, the ambiguity of valid rotations around the vertical axis by multiples of \SI{90}{\degree} needs to be considered.
        To this aim, we iteratively apply
        
        \begin{equation}
            \delta_{h}=\begin{cases}
                \delta_{h} - \SI{90}{\degree} & \delta_{h}\geqslant\SI{45}{\degree} \\
                \delta_{h} & else \\
            \end{cases}
        \end{equation}
        until $\delta_{h}<\SI{45}{\degree}$.
        
        The proposed evaluation metrics $\delta_{v}$ and $\delta_{h}$ can be determined for multiple randomly chosen rotations within the mentioned ranges of $[\SI{-30}{\degree},\SI{30}{\degree}]$ for the horizontal axes and $[\SI{-180}{\degree},\SI{180}{\degree})$ for the vertical axis in sufficient quantity to allow for a statistical analysis.
    \section{Results}
\label{sec:results}

\begin{figure}
    \centering
    \subfigure['Office'.] {
      \includegraphics[width=6cm]{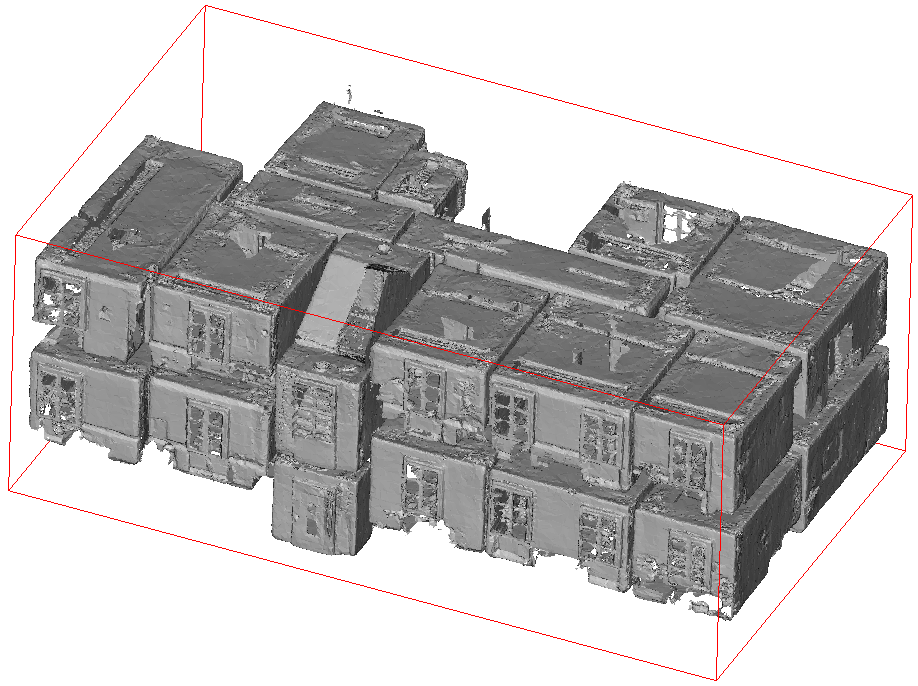}
    }
    \subfigure['Basement'.] {
      \includegraphics[width=6cm]{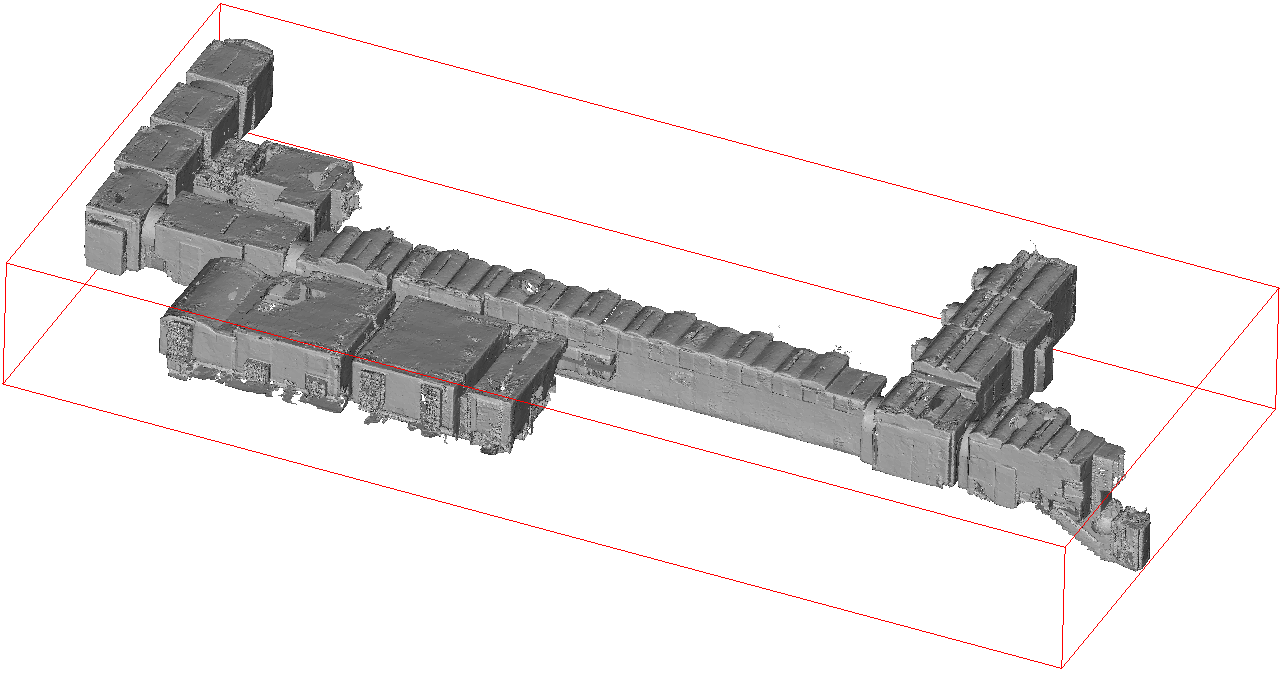}
    }
    \subfigure['Attic.'] {
        \label{fig:results_hololens_attic}
        \includegraphics[width=6cm]{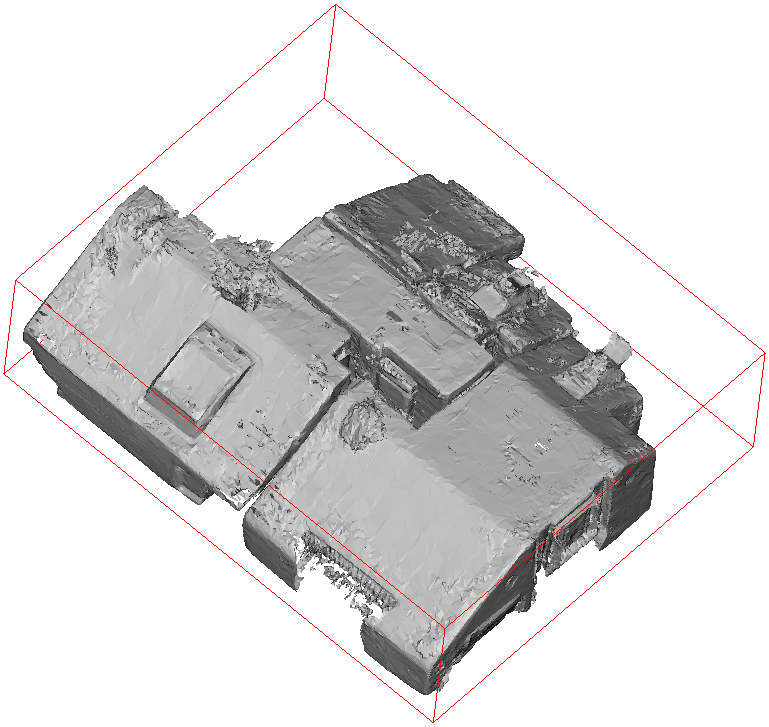}
    }
    \subfigure['Residential House'.] {
      \includegraphics[width=6cm]{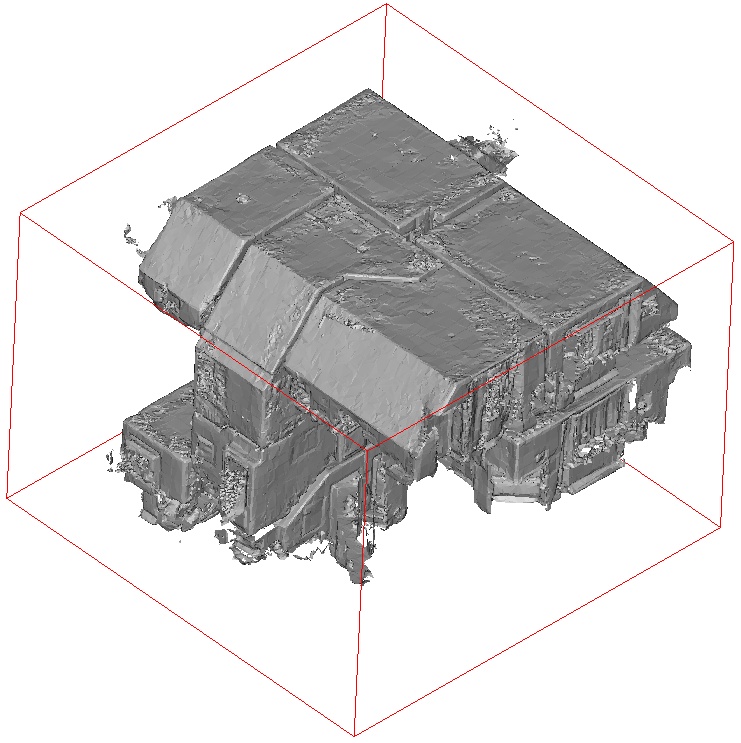}
    }
    \caption{
        The Microsoft HoloLens triangle meshes published in \citep{huebner_et_al_2021} used for evaluation.
        The red box indicates the aligned ground truth pose.
    }
    \label{fig:results_hololens}
\end{figure}

    In order to quantitatively evaluate the approach presented in \autoref{sec:method_rotationAroundVerticalAxis} and \autoref{sec:method_orientationOfVerticalAxis}, the evaluation procedure proposed in \autoref{sec:method_evaluation} was applied to a range of different indoor mapping datasets.
    Firstly, the four triangle meshes of the dataset presented in \citep{huebner_et_al_2021} were used for evaluation.
    These triangle meshes are depicted in \autoref{fig:results_hololens}  along with 3D bounding boxes indicating their respective ground truth pose.
    They were acquired by means of the augmented reality headset Microsoft HoloLens providing coarse triangle meshes of its indoor environment.
    In studies evaluating this device for the use case of indoor mapping, its triangle meshes were found to be accurate in the range of few centimeters in comparison to ground truth data acquired by a terrestrial laser scanner \citep{khoshelham_et_al_2019,huebner_et_al_2019, huebner_et_al_2020}.
    
    The alignment with the coordinate axes of the HoloLens triangle meshes as presented in \citep{huebner_et_al_2021} was found to be inaccurate.
    Actually, in \citep{huebner_et_al_2021}, the presented datasets have been automatically aligned with the coordinate axes by means of an early, inferior version of the approach presented in this paper.
    To enable a reasonable evaluation of the proposed approach on these triangle meshes, ground truth poses were determined by manually aligning the datasets with the coordinate axes.
    The newly aligned datasets along with our implementation of the proposed approach and the evaluation procedure will be made available upon acceptance for publication to allow for reproducibility of the presented evaluation results. 
    
    All four represented indoor environments show a clearly defined Manhattan World structure.
    While the dataset 'Office' has mostly horizontal ceiling surfaces with the exception of the stairwell, the datasets 'Attic' and 'Residential House' have slanted ceiling surfaces.
    The dataset 'Basement' on the other hand shows a range of different barrel-shaped ceilings.
    
\begin{figure}
    \centering
    \subfigure['Case Study 1'.] {
      \includegraphics[width=6cm]{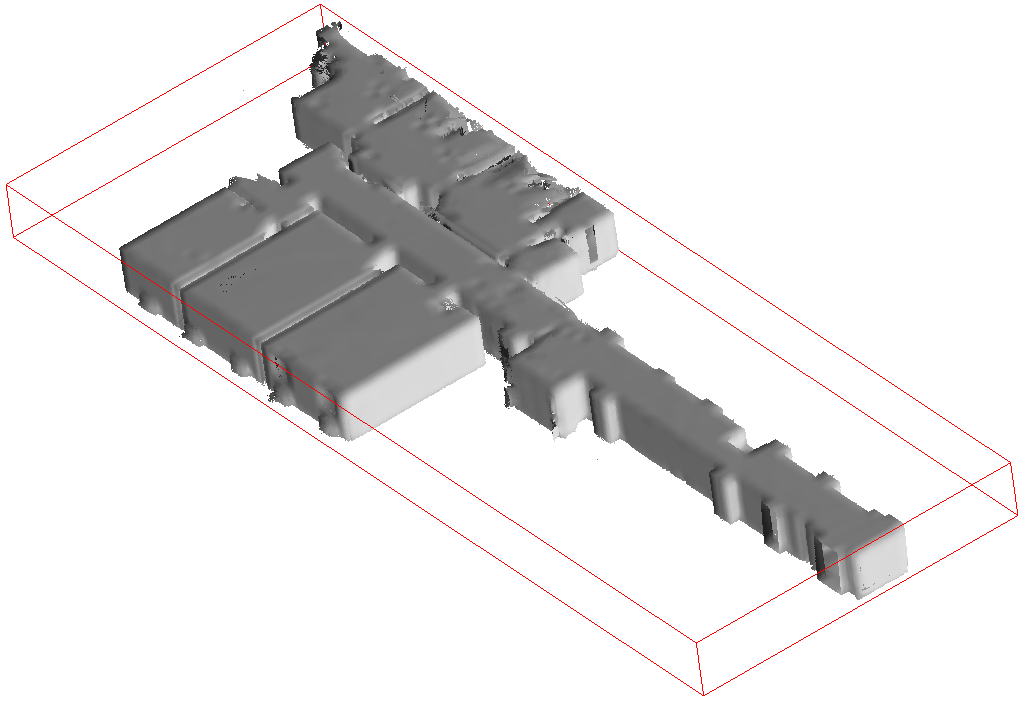}
    }
    \subfigure['Case Study 2'.] {
      \includegraphics[width=6cm]{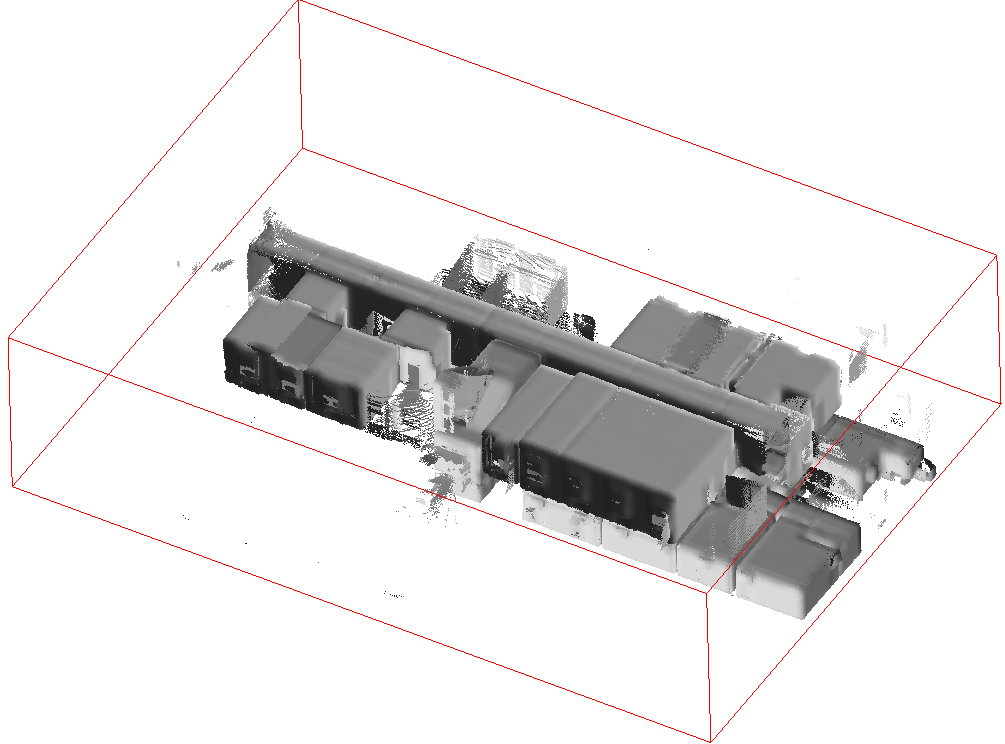}
    }
    \subfigure['Case Study 3'.] {
      \includegraphics[width=6cm]{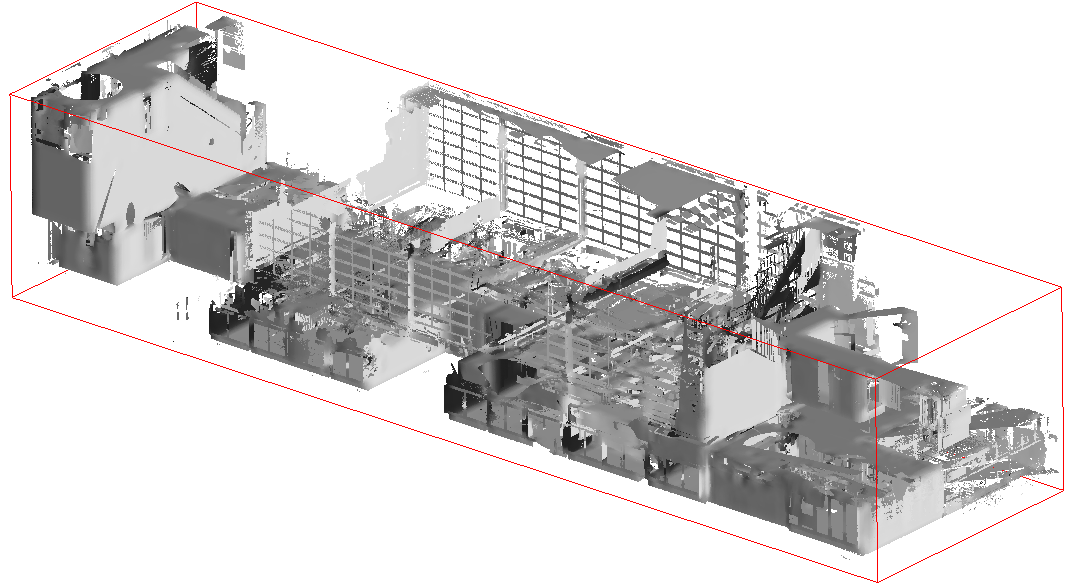}
    }
    \subfigure['Case Study 4'.] {
      \includegraphics[width=6cm]{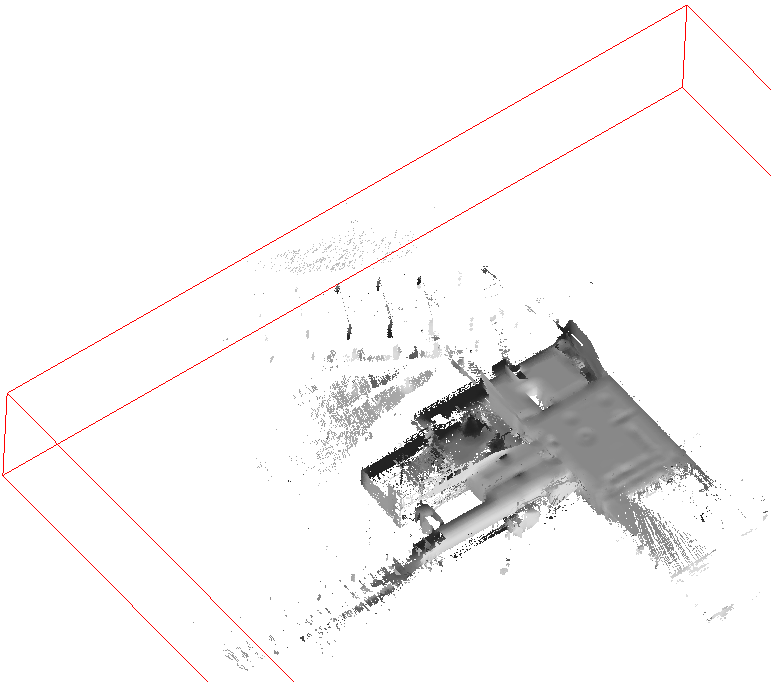}
    }
    \subfigure['Case Study 5'.] {
      \includegraphics[width=6cm]{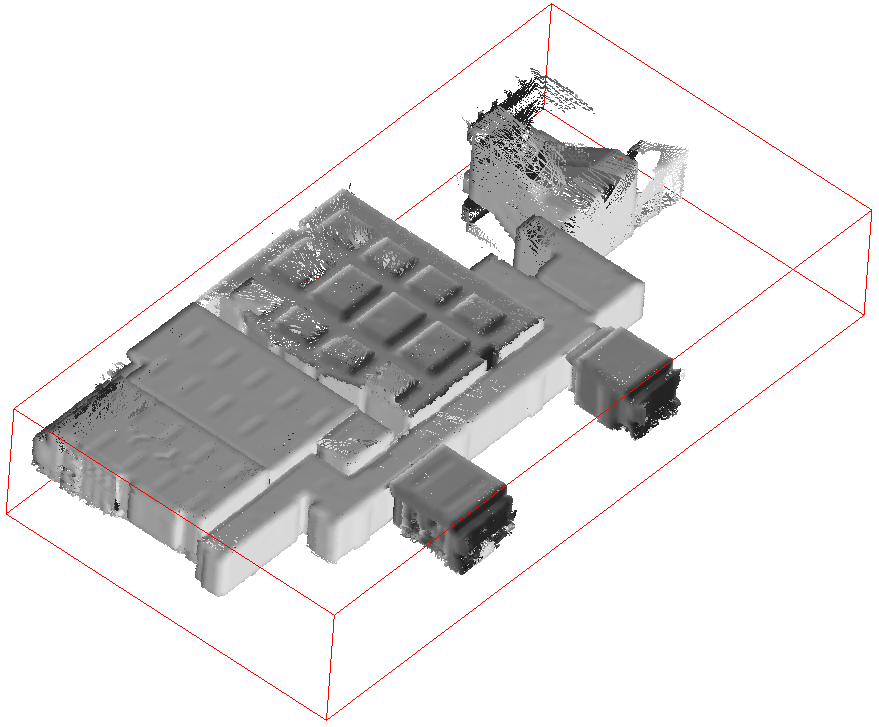}
    }
    \subfigure['Case Study 6'.] {
        \label{fig:results_isprs_caseStudy6}
        \includegraphics[width=6cm]{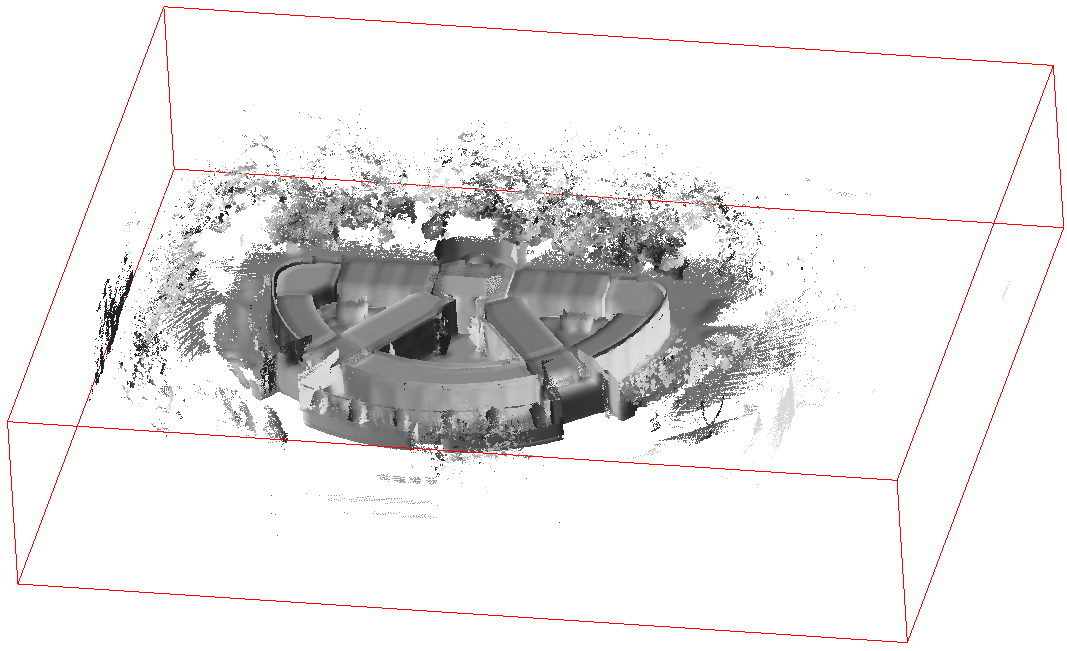}
    }
    \caption{
        The point clouds of the ISPRS Indoor Modelling Benchmark dataset \citep{khoshelham_et_al_2017,khoshelham_et_al_2020} used for evaluation.
        The red box indicates the aligned ground truth pose.
    }
    \label{fig:results_isprs}
\end{figure}

    Furthermore, the six indoor mapping point clouds of the ISPRS Indoor Modelling Benchmark dataset presented in \citep{khoshelham_et_al_2017, khoshelham_et_al_2020} were used for evaluation purposes.
    These point clouds as visualized in \autoref{fig:results_isprs} were acquired by means of different indoor mapping systems with a broad variety of sensor characteristics regarding accuracy and noise.
    Furthermore, the represented indoor environments are characterized by varying amounts of clutter.
    
\begin{figure}
    \centering
    \subfigure[Side view.]{
        \begin{tikzpicture}
            \node at (0, 0) {\includegraphics[width=10cm]{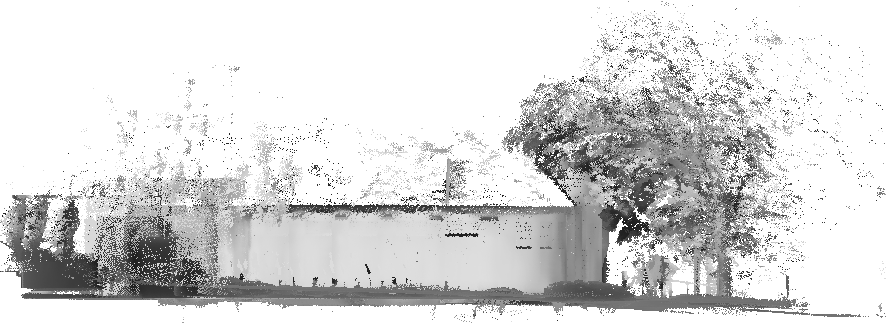}};
            \draw [->, line width=2pt, green] (1.5, -1.0) -- (-1.0, -1.0);
            \draw [->, line width=2pt, blue] (1.5, -1.0) -- (1.5, 1.5);
            \filldraw [red] (1.5, -1.0) circle (0.1);
        \end{tikzpicture}
    }
    \subfigure[Top down view.]{
        \begin{tikzpicture}
            \node at (0, 0) {\includegraphics[width=10cm]{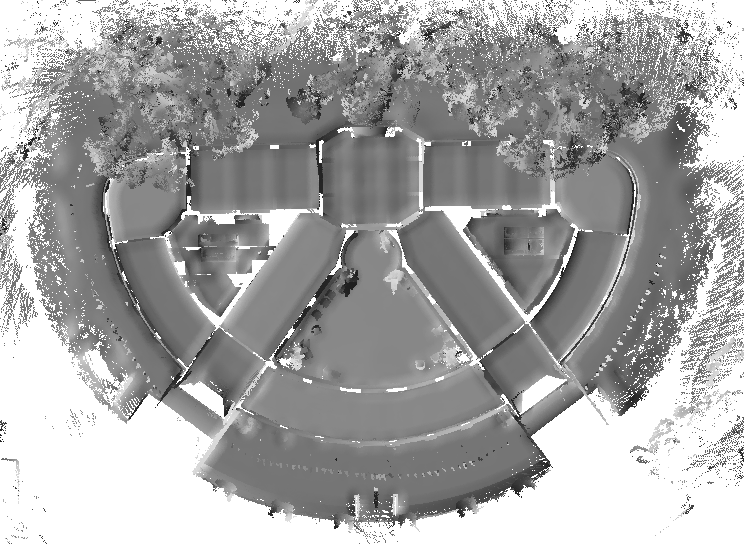}};
            \draw [->, line width=2pt, green] (0, 1.2) -- (0, -1.2);
            \draw [->, line width=2pt, red] (0, 1.2) -- (-2.5, 1.2);
            \filldraw [blue] (0, 1.2) circle (0.1);
        \end{tikzpicture}
    }
    \caption{
        Detailed visualization of the dataset 'Case Study 6' from the ISPRS Indoor Modelling Benchmark dataset \citep{khoshelham_et_al_2020} also depicted in \autoref{fig:results_isprs_caseStudy6}.
        The vertical axis is visualized in blue while the two horizontal axes aligned with the dominant Manhattan World structure of the building are depicted in red and green.
    }
    \label{fig:results_isprs_caseStudy6_details}
\end{figure}
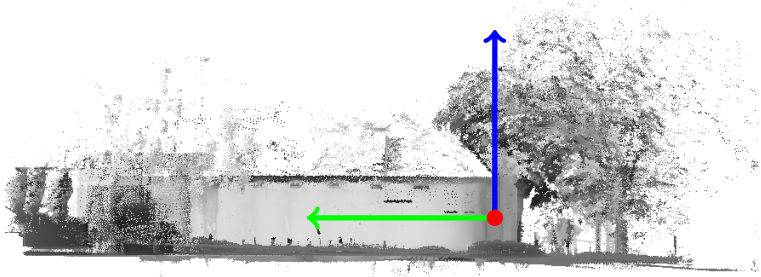
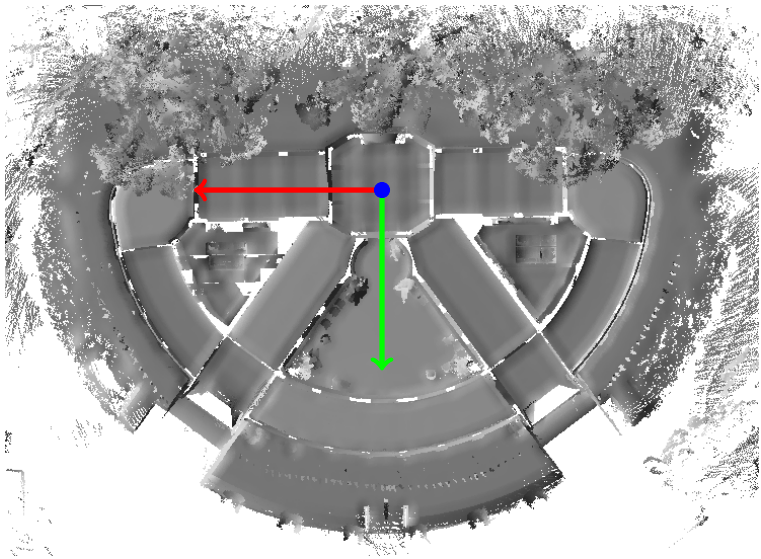

    While the other five datasets mostly adhere to the Manhattan World assumption, the dataset 'Case Study 6' has a high amount of horizontally curved wall surfaces and rooms oriented diagonally with respect to the dominant Manhattan World structure defined by three rooms.
    Furthermore, the point cloud includes a part of the surrounding outdoor terrain with uneven topography and vegetation.
    As the dataset 'Case Study 6' is quite challenging with respect to the aim of this work, it is depicted in more detail in \autoref{fig:results_isprs_caseStudy6_details}.
    
    The point clouds of the ISPRS benchmark dataset as they are published are already aligned with the coordinate axes in accordance with the aim of this work.
    Thus, the poses of the point clouds could directly be used as ground truth poses without any manual adjustment.
    Contrary to triangle meshes however, point clouds do not intrinsically provide normal vectors per point.
    This is also the case with the point clouds of the ISPRS Indoor Modelling Benchmark.
    We thus computed normal vectors for the points after subsampling the point clouds with a resolution of \SI{2}{cm} using CloudCompare 2.10-alpha \cite{cloudcompare}.
    
    Lastly, we also consider some triangle meshes from the Matterport3D dataset \citep{chang_et_al_2017}.
    Matterport3D includes 90 triangle meshes of various kinds of indoor environments acquired with the trolley-mounted Matterport indoor mapping system consisting of multiple RGBD cameras.
    Among the represented indoor environments are some for which the proposed alignment approach is not applicable, as they are not subject to any clearly identifiable Manhattan World structure. 
    Many others do have a clearly identifiable Manhattan World structure but are to a large extent comparable to general building layouts already covered by the HoloLens triangle meshes or ISPRS point clouds used in the scope of this evaluation.
    
\begin{figure}
    \centering
    \subfigure['2azQ1b91cZZ'.] {
      \includegraphics[width=3cm]{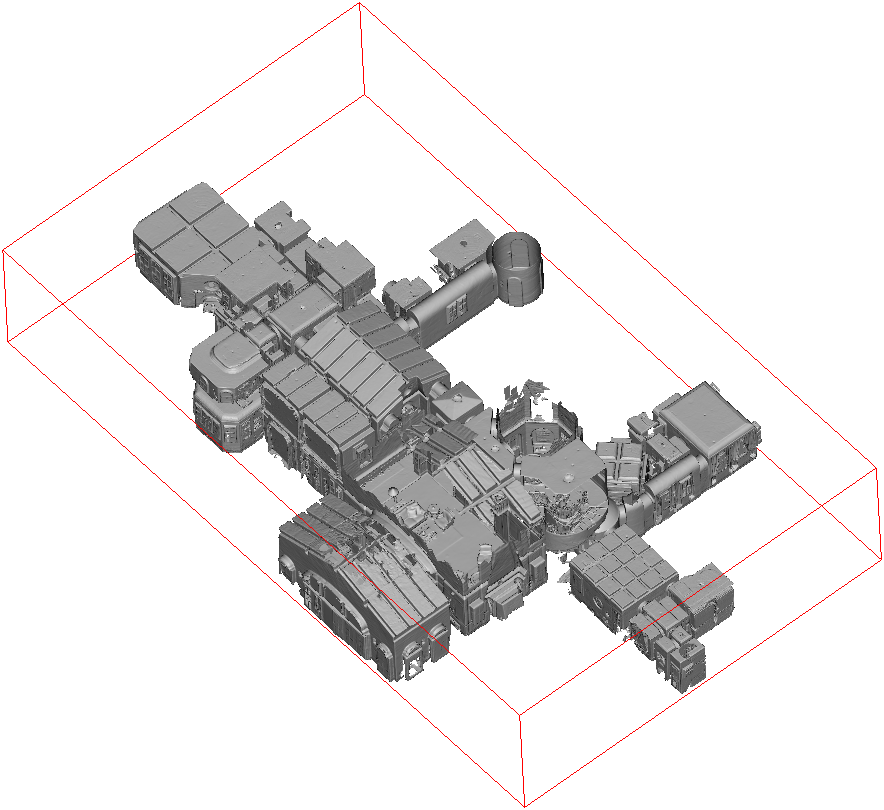}
    }
    \subfigure['759xd9YjKW5'.] {
      \includegraphics[width=3cm]{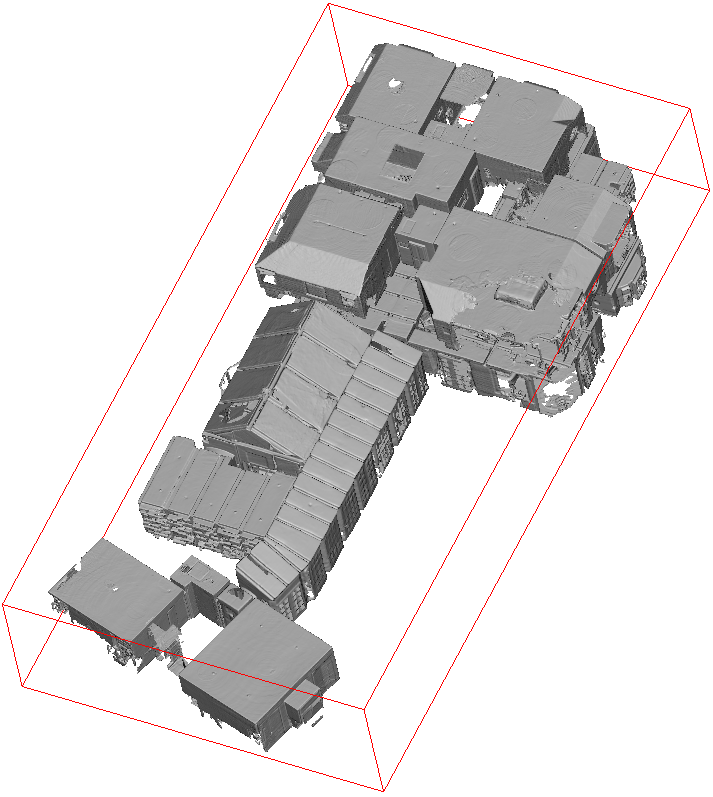}
    }
    \subfigure['ac26ZMwG7aT'.] {
      \includegraphics[width=3cm]{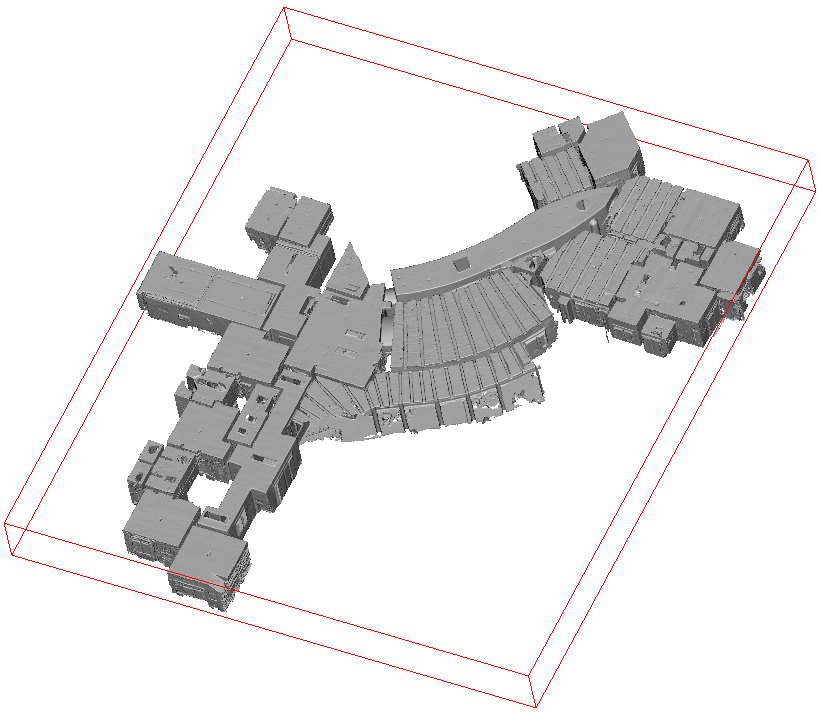}
    }
    \subfigure['fzynW3qQPVF'.] {
      \includegraphics[width=3cm]{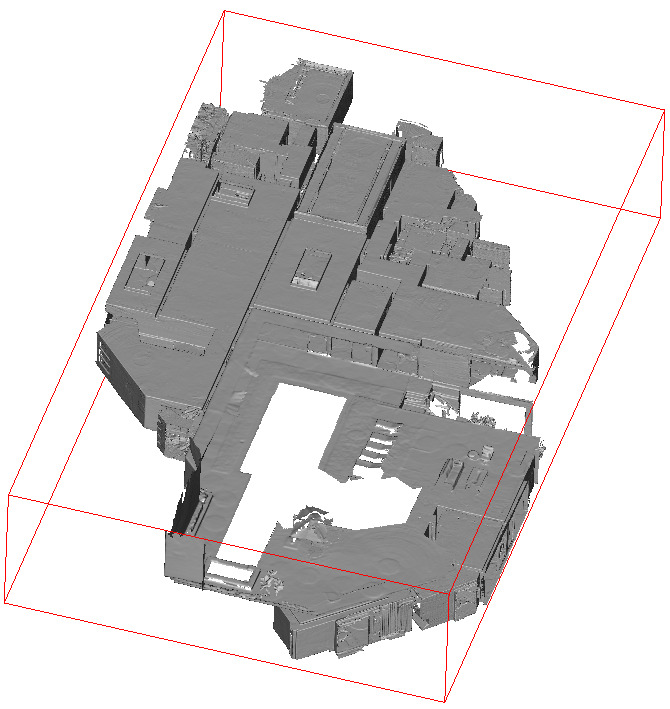}
    }
    \subfigure['gTV8FGcVJC9'.] {
      \includegraphics[width=3cm]{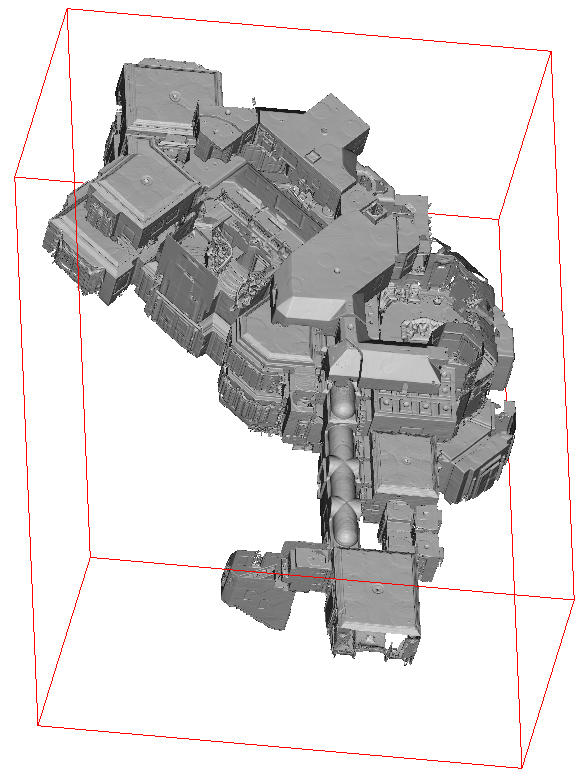}
    }
    \subfigure['mJXqzFtmKg4'.] {
        \label{fig:results_matterport3d_mJXqzFtmKg4}
        \includegraphics[width=3cm]{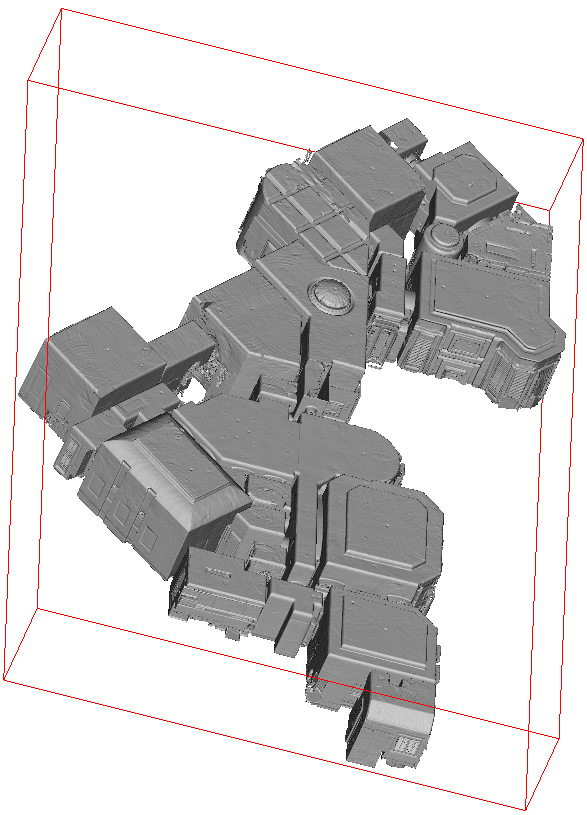}
    }
    \subfigure['p5wJjkQkbXX'.] {
        \includegraphics[width=3cm]{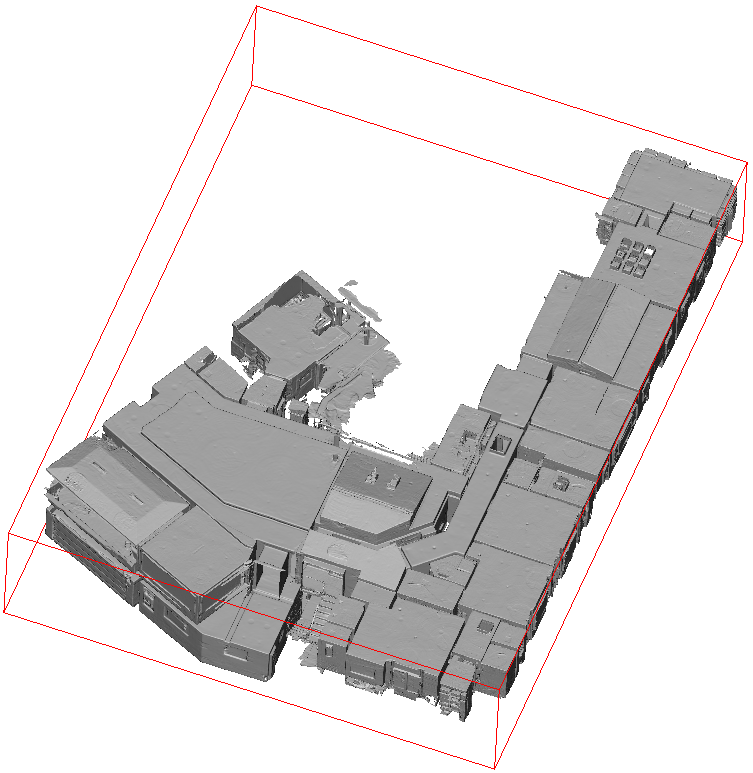}
    }
    \subfigure['PuKPg4mmafe'.] {
        \label{fig:results_matterport3d_PuKPg4mmafe}
        \includegraphics[width=3cm]{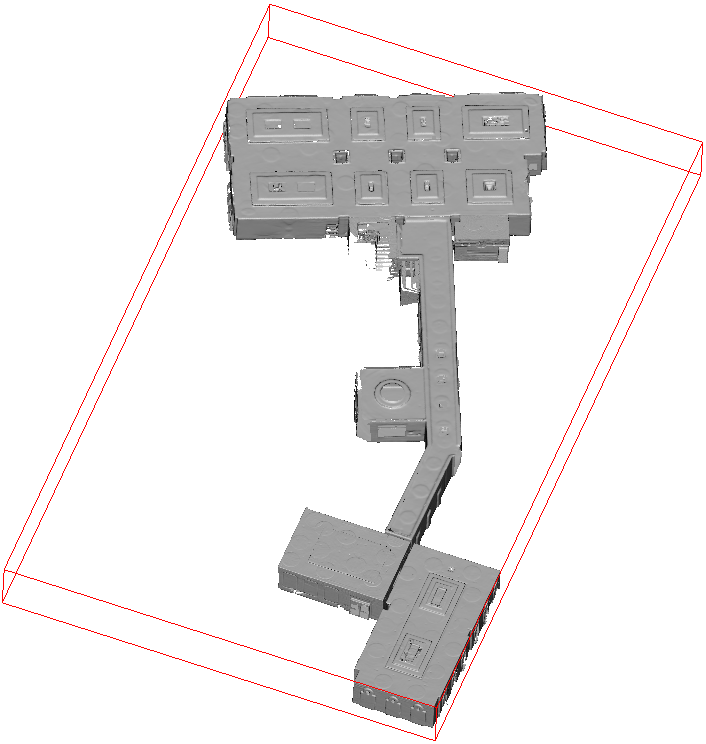}
    }
    \subfigure['ULsKaCPVFJR'.] {
        \label{fig:results_matterport3d_ULsKaCPVFJR}
        \includegraphics[width=3cm]{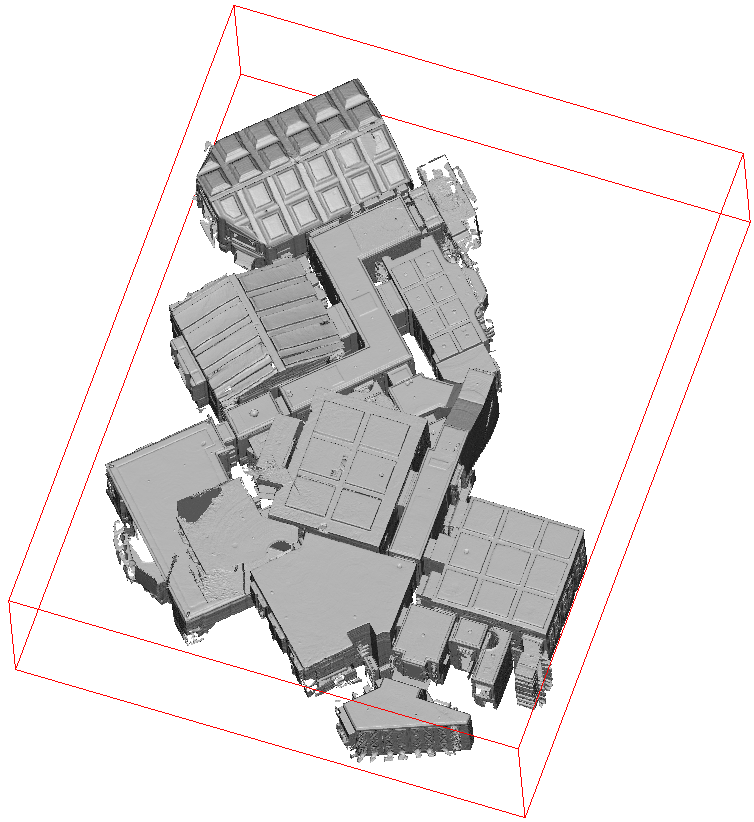}
    }
    \subfigure['ur6pFq6Qu1A'.] {
        \label{fig:results_matterport3d_ur6pFq6Qu1A}
        \includegraphics[width=3cm]{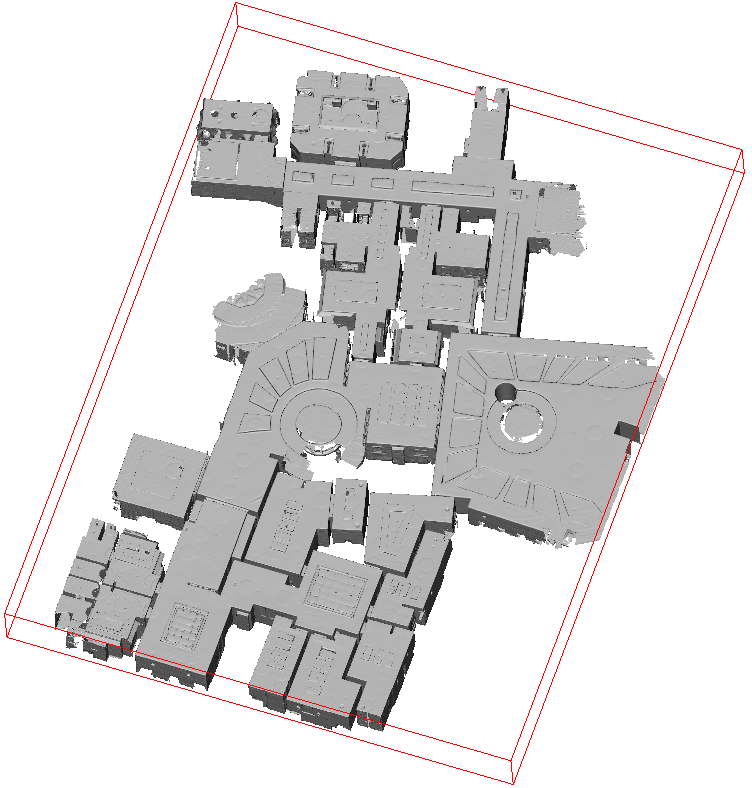}
    }
    \subfigure['VFuaQ6m2Qom'.] {
      \includegraphics[width=3cm]{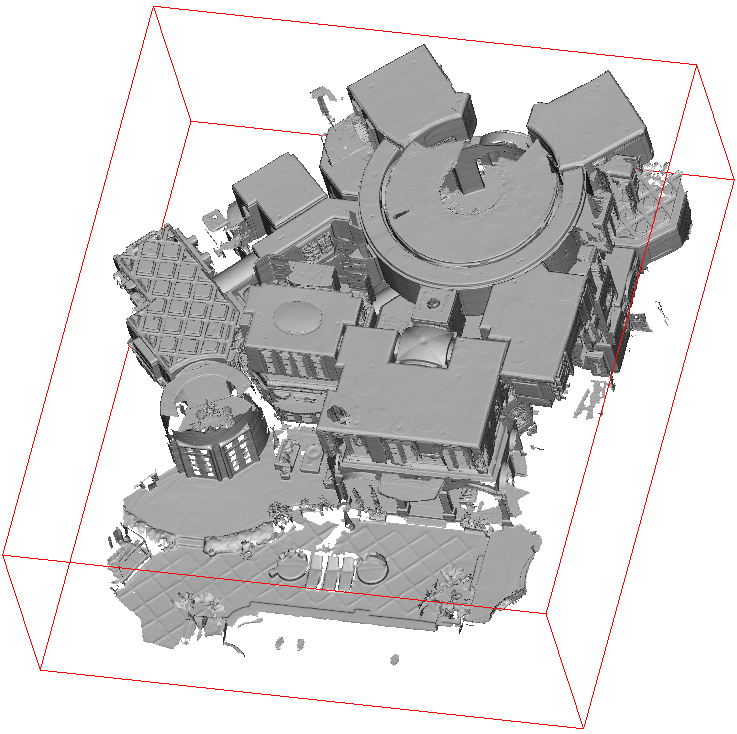}
    }
    \subfigure['Vt2qJdWjCF2'.] {
      \includegraphics[width=3cm]{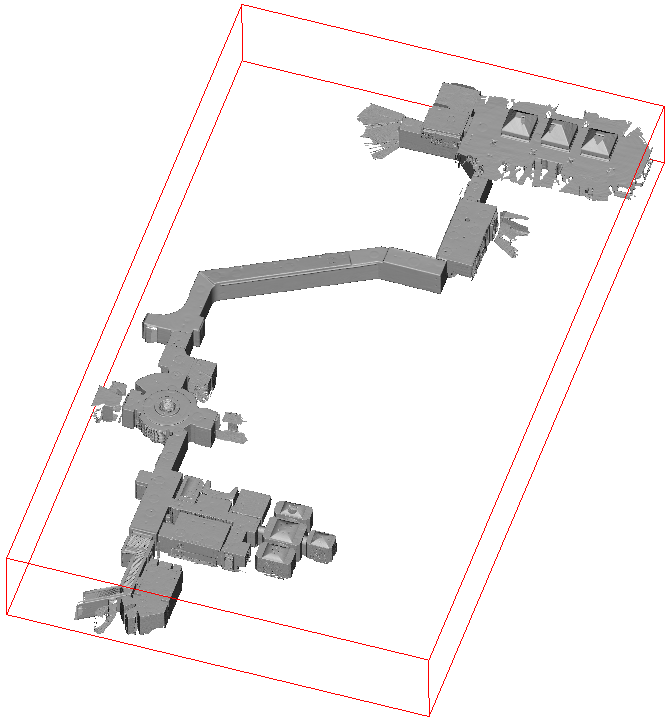}
    }
    \subfigure['x8F5xyUWy9e'.] {
      \includegraphics[width=3cm]{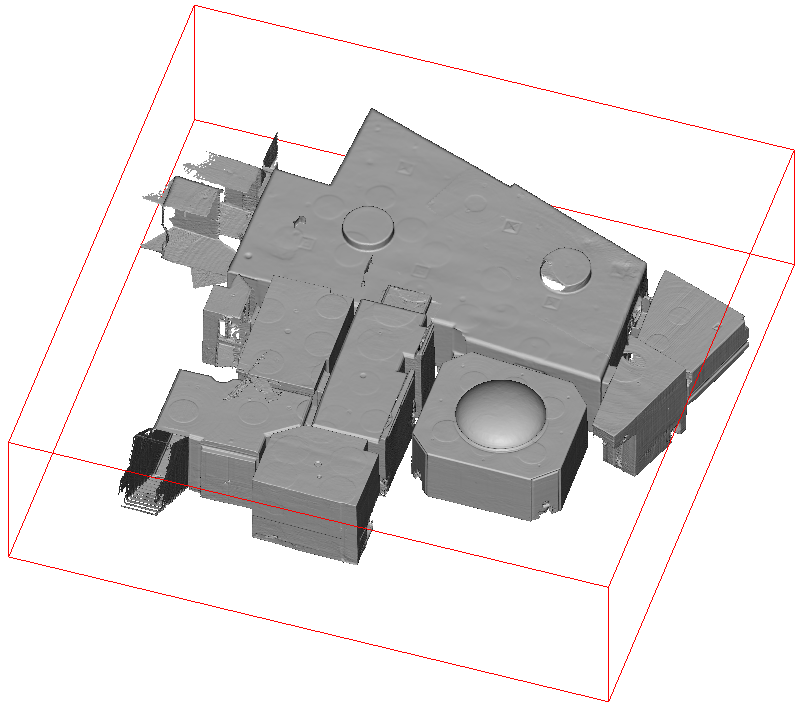}
    }
    \subfigure['ZMojNkEp431'.] {
      \includegraphics[width=3cm]{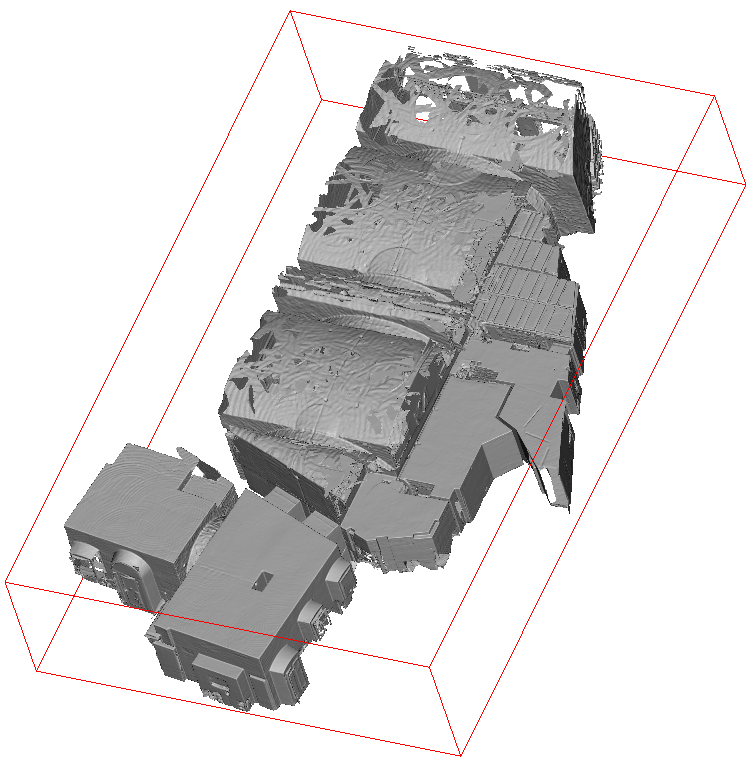}
    }
    \caption{
        The triangle meshes of the Matterport3D dataset \citep{chang_et_al_2017} used for evaluation.
        The red box indicates the aligned ground truth pose.
    }
    \label{fig:results_matterport3d}
\end{figure}

    We thus selected 14 triangle meshes from the Matterport3D dataset that we deem particularly interesting and challenging in the context of this work.
    This, for instance, comprises triangle meshes representing indoor environments that contain more than one underlying Manhattan World system like the one already presented in \autoref{fig:method_rotationAroundVerticalAxis_exampleBuilding}.
    In these cases, the presented alignment method is supposed to align the triangle mesh with the most dominant of the Manhattan World structures at hand being supported by the largest fraction of geometries.
    The 14 selected triangle meshes from the Matterport3D dataset are depicted in \autoref{fig:results_matterport3d}.
    
    As with the ISPRS benchmark point clouds, we again treat the poses of the triangle meshes as they are published as ground truth alignments without any manual adjustments.
    To which extent this decision is justified will be discussed in the subsequent \autoref{sec:discussion}.
    
    The different datasets used in the scope of this evaluation are listed in \autoref{tab:results} along with the respective number of points or triangles comprising them and the respective evaluation results.
    For conducting the evaluation, the evaluation procedure described in \autoref{sec:method_evaluation} was applied to the individual datasets.
    In doing so, each dataset was rotated 50 times while each time, the respective rotation consists of a randomly determined rotation angle $\gamma\in[\SI{-180}{\degree},\SI{180}{\degree})$ around the vertical axis and two random rotations $\alpha,\beta\in[\SI{-30}{\degree},\SI{30}{\degree}]$ around the respective horizontal coordinate axes.
    
    For each of the 50 random input rotations, the alignment procedure described in \autoref{sec:method_rotationAroundVerticalAxis} and \autoref{sec:method_orientationOfVerticalAxis} was applied and the resulting vertical and horizontal angular deviations $\delta_{v}$ and $\delta_{h}$ as defined in \autoref{sec:method_evaluation} were determined.
    \autoref{tab:results} lists mean values and standard deviations for these evaluation metrics aggregated over all 50 samples per dataset.
    Furthermore, mean values and standard deviations for the processing time are given as well.
    The stated values refer to a system with a i7-8550U CPU with \SI{24}{GB} RAM and do not include data import and export.
    The implementation which will be released upon acceptance for publication is CPU-parallelized.
    
    As can be seen in \autoref{tab:results}, the resulting averaged vertical and horizontal angular deviations are largely below \SI{1}{\degree} with the corresponding standard deviations being in a similar range.
    Some outliers marked in red will be discussed in further detail in the subsequent \autoref{sec:discussion}.
    
    \begin{table}[]
    \centering
    \caption{
        Evaluation results for the datasets presented in Figures \ref{fig:results_hololens}, \ref{fig:results_isprs} and \ref{fig:results_matterport3d}.
        The presented values represent 50 randomly chosen orientations per dataset within the range of $[\SI{-180}{\degree},\SI{180}{\degree})$ for rotations around the vertical axis and $[\SI{-30}{\degree},\SI{30}{\degree}]$ for rotations around the horizontal axes.
        The reported numbers of points for the point clouds of the ISPRS Indoor Modelling Benchmark refer to point clouds downsampled to a resolution of \SI{2}{cm} as used in this evaluation.
        The values marked in red are discussed in more detail in \autoref{sec:discussion}.
        \vspace{0.4cm}
    }
    \resizebox{14cm}{!}{
        \begin{tabular}{|c|c|l|r|r|r|r|r|r|r|}
            \hline
            \textbf{Source} & \textbf{Type} & \multicolumn{1}{c|}{\textbf{Dataset}} & \multicolumn{1}{c|}{\textbf{\begin{tabular}[c]{@{}c@{}}Number\\ of \\ Points/\\Triangles\end{tabular}}} & \multicolumn{1}{c|}{\textbf{\begin{tabular}[c]{@{}c@{}}Mean\\  $\delta_{v}$ {[}°{]}\end{tabular}}} & \multicolumn{1}{c|}{\textbf{\begin{tabular}[c]{@{}c@{}}Std.Dev.\\ $\delta_{v}$ {[}°{]}\end{tabular}}} & \multicolumn{1}{c|}{\textbf{\begin{tabular}[c]{@{}c@{}}Mean\\ $\delta_{h}$ {[}°{]}\end{tabular}}} & \multicolumn{1}{c|}{\textbf{\begin{tabular}[c]{@{}c@{}}Std.Dev.\\ $\delta_{h}$ {[}°{]}\end{tabular}}} & \multicolumn{1}{c|}{\textbf{\begin{tabular}[c]{@{}c@{}}Mean\\ Time {[}s{]}\end{tabular}}} & \multicolumn{1}{c|}{\textbf{\begin{tabular}[c]{@{}c@{}}Std.Dev.\\ Time {[}s{]}\end{tabular}}} \\ \hline
            \multirow{4}{*}{\textbf{\begin{tabular}[c]{@{}c@{}}HoloLens \\ \citep{huebner_et_al_2021}\end{tabular}}} & \multirow{4}{*}{\textbf{\begin{tabular}[c]{@{}c@{}}Triangle \\ Mesh\end{tabular}}} & \textbf{Office} & 958,820 & 0.28 & 0.25 & 0.33 & 0.07 & 0.68 & 0.10 \\ \cline{3-3}
             &  & \textbf{Basement} & 695,041 & 0.45 & 0.06 & 0.10 & 0.08 & 0.50 & 0.04 \\ \cline{3-3}
             &  & \textbf{Attic} & 147,146 & \textcolor{red}{3.54} & \textcolor{red}{23.86} & 0.26 & 0.42 & 0.13 & 0.02 \\ \cline{3-3}
             &  & \textbf{Residential House} & 252,820 & 0.16 & 0.05 & 0.71 & 0.42 & 0.30 & 0.04 \\ \hline
            \multirow{6}{*}{\textbf{\begin{tabular}[c]{@{}c@{}}ISPRS \\ Indoor\\ Modelling\\ Benchmark \\ \citep{khoshelham_et_al_2017, khoshelham_et_al_2020}\end{tabular}}} & \multirow{6}{*}{\textbf{\begin{tabular}[c]{@{}c@{}}Point \\ Cloud\end{tabular}}} & \textbf{Case Study 1} & 5,014,452 & 0.01 & 0.05 & 0.03 & 0.16 & 4.41 & 0.19 \\ \cline{3-3}
             &  & \textbf{Case Study 2} & 8,202,319 & 0.01 & 0.02 & 0.01 & 0.13 & 7.40 & 0.26 \\ \cline{3-3}
             &  & \textbf{Case Study 3} & 5,906,718 & 0.02 & 0.01 & 0.04 & 0.17 & 5.68 & 0.29 \\ \cline{3-3}
             &  & \textbf{Case Study 4} & 4,846,736 & 0.01 & 0.26 & 0.03 & 0.44 & 4.19 & 0.27 \\ \cline{3-3}
             &  & \textbf{Case Study 5} & 4,409,794 & 0.02 & 0.07 & 0.02 & 0.06 & 3.96 & 0.23 \\ \cline{3-3}
             &  & \textbf{Case Study 6} & 11,760,325 & 0.02 & 0.02 & 0.06 & 0.77 & 8.65 & 0.53 \\ \hline
            \multirow{14}{*}{\begin{tabular}[c]{@{}c@{}}\textbf{Matterport3D} \\ \citep{chang_et_al_2017}\end{tabular}} & \multirow{14}{*}{\textbf{\begin{tabular}[c]{@{}c@{}}Triangle\\ Mesh\end{tabular}}} & \textbf{2azQ1b91cZZ} & 9,549,830 & 0.03 & 0.02 & 0.44 & 0.06 & 8.24 & 0.39 \\ \cline{3-3}
             &  & \textbf{759xd9YjKW5} & 6,208,440 & 0.05 & 0.01 & 0.18 & 0.05 & 5.48 & 0.35 \\ \cline{3-3}
             &  & \textbf{ac26ZMwG7aT} & 10,811,581 & 0.05 & 0.09 & 0.52 & 0.06 & 9.84 & 0.49 \\ \cline{3-3}
             &  & \textbf{fzynW3qQPVF} & 9,105,979 & 0.09 & 0.02 & 0.05 & 0.06 & 10.75 & 0.60 \\ \cline{3-3}
             &  & \textbf{gTV8FGcVJC9} & 14,436,867 & 0.05 & 0.05 & 0.11 & 0.07 & 12.29 & 0.96 \\ \cline{3-3}
             &  & \textbf{mJXqzFtmKg4} & 8,237,802 & 0.07 & 0.33 & \textcolor{red}{2.73} & \textcolor{red}{14.29} & 6.90 & 0.54 \\ \cline{3-3}
             &  & \textbf{p5wJjkQkbXX} & 10,678,539 & 0.07 & 0.02 & 0.40 & 0.03 & 10.35 & 0.68 \\ \cline{3-3}
             &  & \textbf{PuKPg4mmafe} & 1,968,102 & 0.05 & 0.01 & \textcolor{red}{15.28} & \textcolor{red}{20.07} & 1.83 & 0.11 \\ \cline{3-3}
             &  & \textbf{ULsKaCPVFJR} & 6,612,194 & 0.05 & 0.01 & \textcolor{red}{44.41} & 0.04 & 5.51 & 0.47 \\ \cline{3-3}
             &  & \textbf{ur6pFq6Qu1A} & 9,277,187 & 0.02 & 0.01 & \textcolor{red}{12.85} & 0.05 & 9.42 & 0.42 \\ \cline{3-3}
             &  & \textbf{VFuaQ6m2Qom} & 9,453,891 & 0.03 & 0.02 & 0.13 & 0.06 & 8.53 & 0.37 \\ \cline{3-3}
             &  & \textbf{Vt2qJdWjCF2} & 6,429,106 & 0.10 & 0.01 & 0.05 & 0.09 & 6.40 & 0.38 \\ \cline{3-3}
             &  & \textbf{x8F5xyUWy9e} & 2,862,858 & 0.07 & 0.01 & 0.21 & 0.08 & 2.66 & 0.16 \\ \cline{3-3}
             &  & \textbf{ZMojNkEp431} & 4,690,777 & 0.06 & 0.05 & 0.18 & 0.08 & 4.31 & 0.27 \\ \hline
        \end{tabular}
    }
    \label{tab:results}
\end{table}
    \section{Discussion}
\label{sec:discussion}
    
    Taking a closer look at the evaluation results presented in \autoref{tab:results}, the overall quite low values for the horizontal and vertical angular deviations $\delta_{h}$ and $\delta_{v}$ with overall equally low standard deviations indicate that the proposed alignment method works overall quite well for a large range of different indoor mapping point clouds and triangle meshes with randomly varying input rotations within the defined bounds.
    The consistently larger $\delta_{v}$ and $\delta_{h}$ values for the triangle meshes acquired with the Microsoft HoloLens may be attributable to them being less accurate and more affected by noise.
    Triangles pertaining to an actually smooth planar room surface show a considerable variation in normal vector direction.
    However, the reported $\delta_{v}$ and $\delta_{h}$ values for these datasets are still mostly well below \SI{1}{\degree}.
    
    Some datasets however show significantly higher averaged values for $\delta_{v}$ or $\delta_{h}$, sometimes with the corresponding standard variation being significantly raised as well.
    These outliers are marked red in \autoref{tab:results} and will be discussed in more detail in the following paragraphs.
    To analyze these cases, we will take a closer look at the distribution of the individual 50 deviations constituting the respective mean value and standard deviation.
    
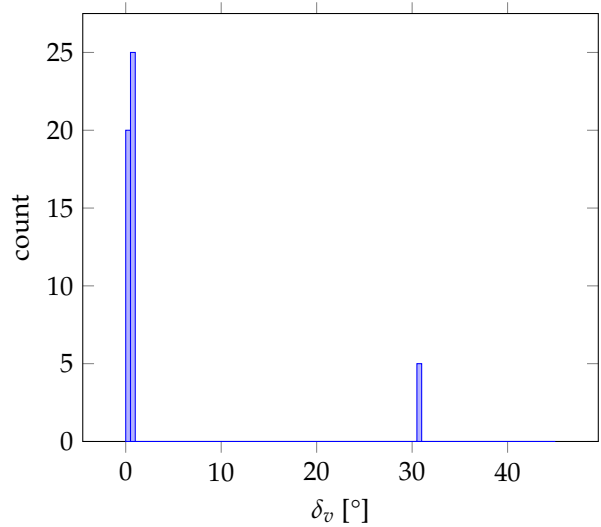
\begin{figure}
    \centering
    \begin{tikzpicture}
        \begin{axis}[
            ybar, 
            ymin=0,
            xlabel near ticks,
            ylabel near ticks,
            xlabel={$\delta_{v}$ {[}\textdegree{]}},
            ylabel={count}
        ]
            \addplot +[
                hist={
                    bins=90,
                    data min=0,
                    data max=45
                }   
            ] table [y index=0] {figures/data/results_hololens_attic_dv.csv};
        \end{axis}
    \end{tikzpicture}
    \caption{
        Histogram of the 50 $\delta_{v}$ values resulting in the mean value of $\SI{3.54}{\degree}\pm\SI{23.86}{\degree}$ presented in \autoref{tab:results} for the triangle mesh 'Attic' depicted in \autoref{fig:results_hololens_attic}.
        Without the 5 outliers around \SI{31}{\degree}, mean $\delta_{v}$ results in $\SI{0.50}{\degree}\pm\SI{0.13}{\degree}$.
    }
    \label{fig:discussion_hololens_attic_dv_histogram}
\end{figure}

\begin{figure}
    \centering
    \begin{tikzpicture}
        \node at (0,0){         
            \includegraphics[width=10cm]{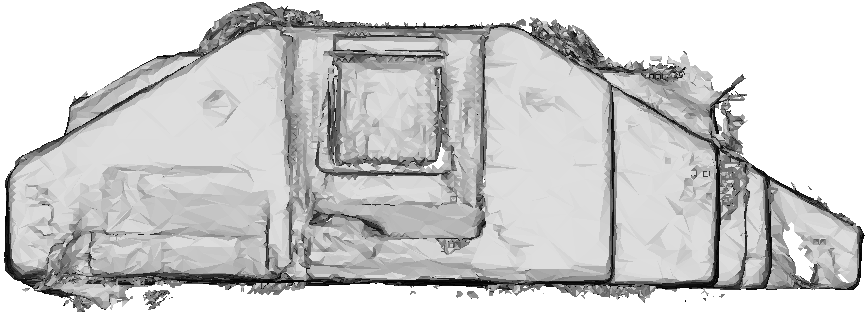}
        };
        \draw [draw=green, line width=1pt] (-5.0,-1.9) rectangle (5.0,1.9);
        \begin{scope}[rotate around={30.94:(0,0)}]
            \draw [draw=red, line width=1pt] (-4.95,-3.85) rectangle (3.9,2.75);    
        \end{scope}
    \end{tikzpicture}
    \caption{
        Resulting vertical alignments of the triangle mesh 'Attic' from \autoref{fig:results_hololens_attic} for the two peaks in the histogram of $\delta_{v}$ values depicted in \autoref{fig:discussion_hololens_attic_dv_histogram}.
        The green bounding box corresponds to the peak at $\delta_{v}\approx\SI{0}{\degree}$ while the red bounding box corresponds to the minor peak at $\delta_{v}\approx\SI{30}{\degree}$.
    }
    \label{fig:discussion_hololens_attic}
\end{figure}

    In the case of the HoloLens triangle mesh 'Attic' for instance, the histogram of $\delta_{v}$ values depicted in \autoref{fig:discussion_hololens_attic_dv_histogram} indicates that the heightened mean and standard deviation values for the angular deviation in the vertical alignment are not caused by a large variability in the resulting vertical alignment.
    The vertical orientations resulting from the evaluated alignment method rather fluctuate between two clearly defined states, one being the correct vertical orientation according to the ground truth pose at around \SI{0}{\degree} angular deviation $\delta_{v}$ of the vertical axis supported by 45 of the 50 measurements.
    The other state is a vertical orientation with an angular deviation of about \SI{30}{\degree} occurring in the remaining five measurements.
    As visualized by the red box in \autoref{fig:discussion_hololens_attic}, this corresponds to an alignment where the vertical axis is oriented orthogonally to one of the slanted ceiling surfaces.
    
    This is the only case where the vertical alignment did not work satisfyingly in all 50 samples for all the datasets used in the evaluation.
    We suspect that the misalignments occurring sporadically on this dataset can be ascribed to the noisy surfaces of the HoloLens triangle meshes.
    The triangles comprising the large horizontal floor surface for instance differ significantly in the direction of their normal vectors.
    Thus, only a fraction of the triangles comprising the floor actually corresponds to the proper vertical direction with respect to the applied resolution of \SI{1}{\degree}.
    Depending on the input rotation, a peak caused by a slanted ceiling surface with a not insignificant area in comparison to horizontal surfaces like in the case of the dataset at hand representing only the attic story may thus induce a larger peak and consequently a misalignment.
    In cases like this, applying an angular resolution of more than \SI{1}{\degree} may be more suited to prevent suchlike misalignments.
    
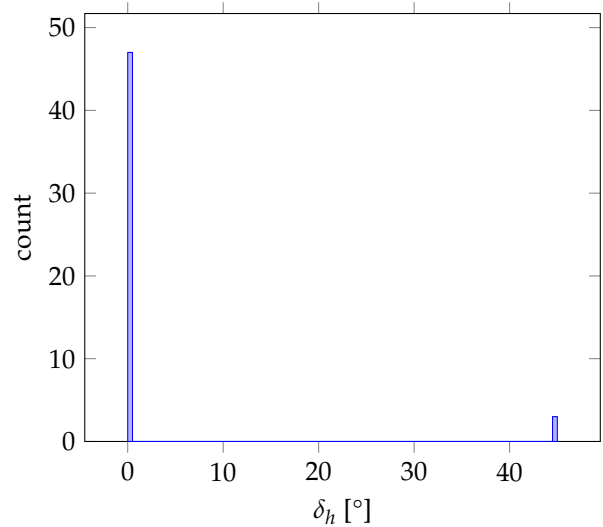
\begin{figure}
    \centering
    \begin{tikzpicture}
        \begin{axis}[
            ybar, 
            ymin=0,
            xlabel near ticks,
            ylabel near ticks,
            xlabel={$\delta_{h}$ {[}\textdegree{]}},
            ylabel={count}
        ]
            \addplot +[
                hist={
                    bins=90,
                    data min=0,
                    data max=45
                }   
            ] table [y index=0] {figures/data/results_hololens_matterport3d_mJXqzFtmKg4_dh.csv};
        \end{axis}
    \end{tikzpicture}
    \caption{
        Histogram of the 50 $\delta_{h}$ values resulting in the mean value of $\SI{2.73}{\degree}\pm\SI{14.29}{\degree}$ presented in \autoref{tab:results} for the triangle mesh 'mJXqzFtmKg4' depicted in \autoref{fig:results_matterport3d_mJXqzFtmKg4}.
    }
    \label{fig:discussion_matterport3d_mJXqzFtmKg4_dh_histogram}
\end{figure}

\begin{figure}
    \centering
    \begin{tikzpicture}
        \node at (0,0){         
            \includegraphics[width=10cm]{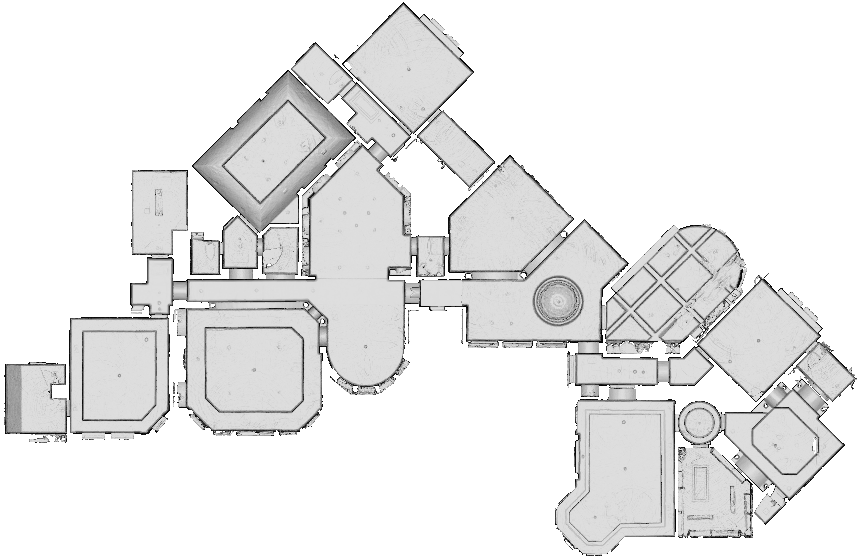}
        };
        \draw [draw=green, line width=1pt] (-5.0,-3.25) rectangle (5.0,3.25);
        \begin{scope}[rotate around={44.8:(0,0)}]
            \draw [draw=red, line width=1pt] (-4.8,-4.77) rectangle (2.9,3.35);    
        \end{scope}
    \end{tikzpicture}    
    \caption{
        Resulting horizontal alignments of the triangle mesh 'mJXqzFtmKg4' from \autoref{fig:results_matterport3d_mJXqzFtmKg4} for the two peaks in the histogram of $\delta_{h}$ values depicted in \autoref{fig:discussion_matterport3d_mJXqzFtmKg4_dh_histogram}.
        The green bounding box corresponds to the peak at $\delta_{h}\approx\SI{0}{\degree}$ while the red bounding box corresponds to the minor peak at $\delta_{h}\approx\SI{45}{\degree}$.
    }
    \label{fig:discussion_matterport3d_mJXqzFtmKg4}
\end{figure}

    Besides the discussed outlier in the vertical aligment, some outliers in the horizontal alignment do exist.
    The Matterport3D datasets 'mJXqzFtmKg4' and 'PuKPg4mmafe' for instance show heightened average $\delta_{h}$ values along with high standard deviations.
    The histograms showing the distribution of all 50 $\delta_{h}$ values are again depicted in \autoref{fig:discussion_matterport3d_mJXqzFtmKg4_dh_histogram} and \autoref{fig:discussion_matterport3d_PuKPg4mmafe_dh_histogram} respectively.
    Like in the case before, it is apparent that the alignment results fluctuate between two states depending on the input rotation for both cases while each time, one peak at \SI{0}{\degree} corresponds to the correct horizontal alignment according to the respective ground truth pose.
    As can be seen in \autoref{fig:discussion_matterport3d_mJXqzFtmKg4} and \autoref{fig:discussion_matterport3d_PuKPg4mmafe}, the respective second peak corresponds in both cases to a valid second Manhattan World structure present in the respective indoor environment.
    
    \begin{figure}
    \centering
    \begin{tikzpicture}
        \begin{axis}[
            ybar, 
            ymin=0,
            xlabel near ticks,
            ylabel near ticks,
            xlabel={$\delta_{h}$ {[}\textdegree{]}},
            ylabel={count}
        ]
            \addplot +[
                hist={
                    bins=90,
                    data min=0,
                    data max=45
                }   
            ] table [y index=0] {figures/data/results_hololens_matterport3d_PuKPg4mmafe_dh.csv};
        \end{axis}
    \end{tikzpicture}
    \caption{
        Histogram of the 50 $\delta_{h}$ values resulting in the mean value of $\SI{15.28}{\degree}\pm\SI{20.07}{\degree}$ presented in \autoref{tab:results} for the triangle mesh 'PuKPg4mmafe' depicted in \autoref{fig:results_matterport3d_PuKPg4mmafe}.
    }
    \label{fig:discussion_matterport3d_PuKPg4mmafe_dh_histogram}
\end{figure}
    
    \begin{figure}
    \centering
    \begin{tikzpicture}
        \node at (0,0){         
            \includegraphics[width=10cm]{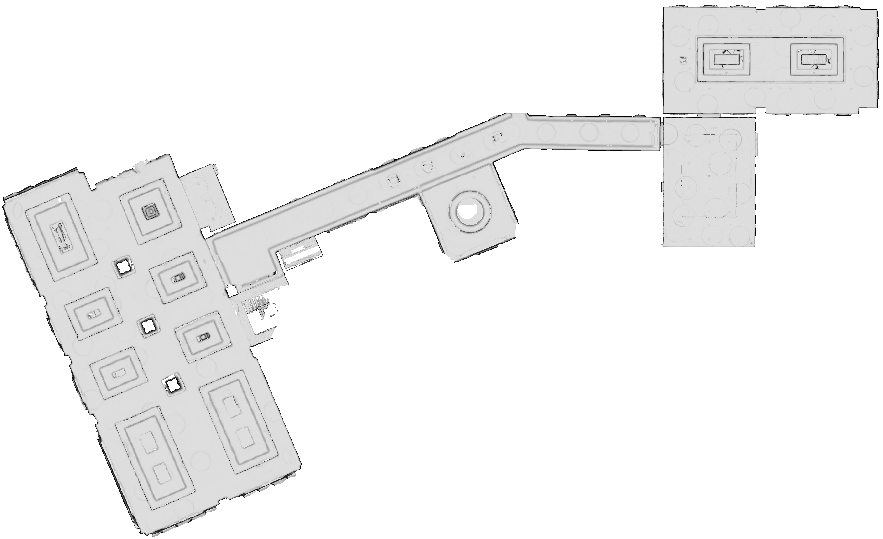}
        };
        \draw [draw=green, line width=1pt] (-5.05,-3.1) rectangle (5.05,3.1);
        \begin{scope}[rotate around={23.0:(0,0)}]
            \draw [draw=red, line width=1pt] (-4.4,-1.55) rectangle (5.75,2.75);    
        \end{scope}
    \end{tikzpicture}
    \caption{
        Resulting horizontal alignments of the triangle mesh 'PuKPg4mmafe' from \autoref{fig:results_matterport3d_PuKPg4mmafe} for the two peaks in the histogram of $\delta_{h}$ values depicted in \autoref{fig:discussion_matterport3d_PuKPg4mmafe_dh_histogram}.
        The green bounding box corresponds to the peak at $\delta_{v}\approx\SI{0}{\degree}$ while the red bounding box corresponds to the peak at $\delta_{v}\approx\SI{23}{\degree}$.
    }
    \label{fig:discussion_matterport3d_PuKPg4mmafe}
\end{figure}
    
    In the case of the dataset 'mJXqzFtmKg4', this seems immediately plausible, as both Manhattan World structures present in the indoor environment are supported by a comparable amount of geometries, as was already demonstrated in \autoref{fig:method_rotationAroundVerticalAxis_grid} and \autoref{fig:method_rotationAroundVerticalAxis_horizontalFaces}.
    Thus, different input rotations may result in slightly different discretizations within the grid of \SI{1}{\degree} resolution, sometimes favoring one and sometimes the other Manhattan World structure as having the largest peak of summarized geometry weights.
    
    In the case of the dataset 'PuKPg4mmafe' however, the two Manhattan World structures present in the indoor environment apparently do not seem to be supported by an approximately equal fraction of geometries.
    Rather, the upper right section in \autoref{fig:discussion_matterport3d_PuKPg4mmafe} constituting the one Manhattan World structure seems to be far smaller than the section on the lower left constituting the other Manhattan World structure.
    In this case, the ground truth pose of the triangle mesh as published in \citep{chang_et_al_2017} is aligned with the apparently smaller Manhattan World structure.
    It is thus not surprising that in the evaluation, a majority of measurements results in high $\delta_{h}$ deviations as the evaluated alignment method favors the larger Manhattan World structure.
    However, it is surprising that a not insignificant fraction of 17 of the 50 randomly chosen input rotations results in a horizontal alignment along the apparently significantly smaller Manhattan World structure.
    
\begin{figure}
    \centering
    \includegraphics[width=12cm]{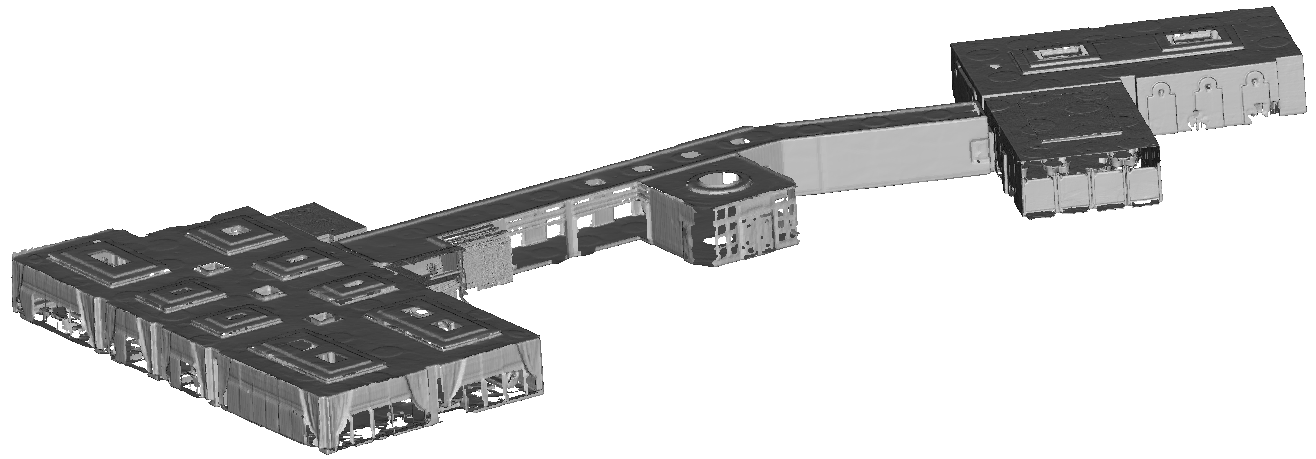}
    \caption{
        Detailed view of the triangle mesh 'PuKPg4mmafe' from the Matterport3D dataset also depicted in \autoref{fig:results_matterport3d_PuKPg4mmafe} and \autoref{fig:discussion_matterport3d_PuKPg4mmafe}.
        Note that in the case of the larger part of the building structure determining the Manhattan World system visualized by the red bounding box in \autoref{fig:discussion_matterport3d_PuKPg4mmafe}, large parts of the wall surfaces are missing as wall openings or constituted by curtains or other structures with inhomogeneous normal direction.
        The smaller part of the building structure on the right hand side which determines the Manhattan World system visualized by the green bounding box in \autoref{fig:discussion_matterport3d_PuKPg4mmafe} however has largely closed, smooth wall surfaces.
    }
    \label{fig:discussion_matterport3d_PuKPg4mmafe_2}
\end{figure}

    This situation may be explainable by taking a closer look at the walls constituting the respective Manhattan World structures.
    As can be seen in \autoref{fig:discussion_matterport3d_PuKPg4mmafe_2}, the smaller Manhattan World section on the right hand side consists of wall surfaces that are generally smooth and completely covered with geometries.
    The larger section on the left however has a large fraction of open wall surface were there are no geometries due to the walls there actually being openings or glass surfaces that cannot be captured by the Matterport system used for the acquisition of this dataset.
    Furthermore, large parts of the actually represented wall surfaces are covered with curtains or other structures resulting in inhomogeneous normal vector directions.
    In consideration of this, it seems plausible that the actual support for both Manhattan World structures present in the building could be approximately equal and the applied alignment method could thus be prone to fluctuate between both Manhattan World systems with varying input rotations.
    
    Besides these both cases discussed so far, there are two further datasets with high average horizontal angular alignment deviations in the evaluation results reported in \autoref{tab:results}.
    These are the triangle meshes 'ULsKaCPVFJR' and 'ur6pFq6Qu1A' which are also part of the Matterport3D dataset.
    Unlike the cases discussed before, these however only show heightened mean values for $\delta_{h}$ while the respective standard deviations are low in a range comparable to the other Matterport3D triangle meshes where the evaluated alignment method proofed to be consistently successful.
    
\begin{figure}
    \centering
    \begin{tikzpicture}
        \node at (0,0){         
            \includegraphics[width=10cm]{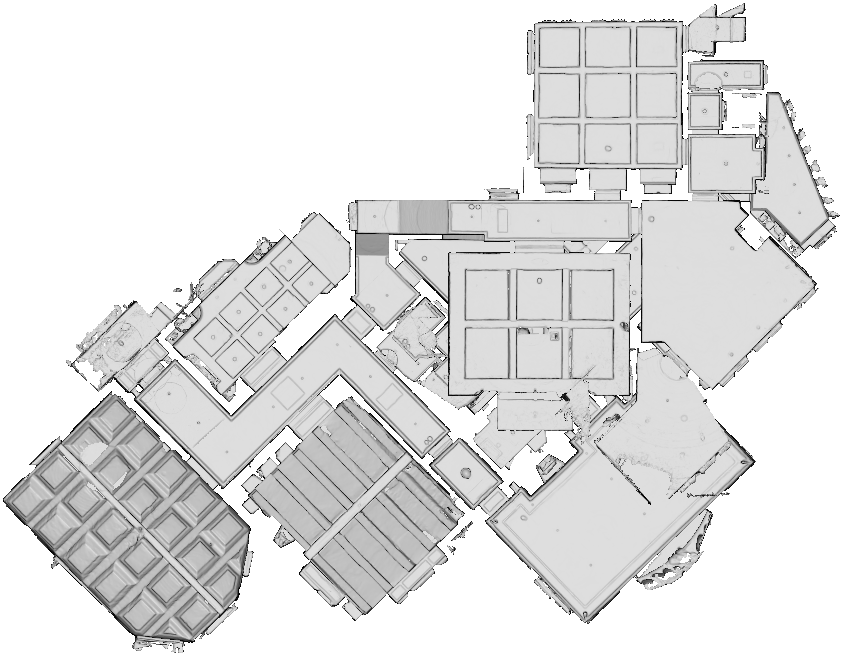}
        };
        \draw [draw=green, line width=1pt] (-5.05,-3.9) rectangle (5.05,3.9);
        \begin{scope}[rotate around={44.41:(0,0)}]
            \draw [draw=red, line width=1pt] (-5.15,-4.3) rectangle (5.47,2.8);    
        \end{scope}
    \end{tikzpicture}
    \caption{
        The green bounding box represents the horizontal alignment of the triangle mesh 'ULsKaCPVFJR' from \autoref{fig:results_matterport3d_ULsKaCPVFJR} as it is published in \citep{chang_et_al_2017} and used as ground truth pose for the evaluation results presented in  \autoref{tab:results}.
        The red bounding box on the other hand represents the horizontal alignment resulting from our presented approach.
    }
    \label{fig:discussion_matterport3d_ULsKaCPVFJR}
\end{figure}

\begin{figure}
    \centering
    \begin{tikzpicture}
        \node at (0,0){         
            \includegraphics[width=10cm]{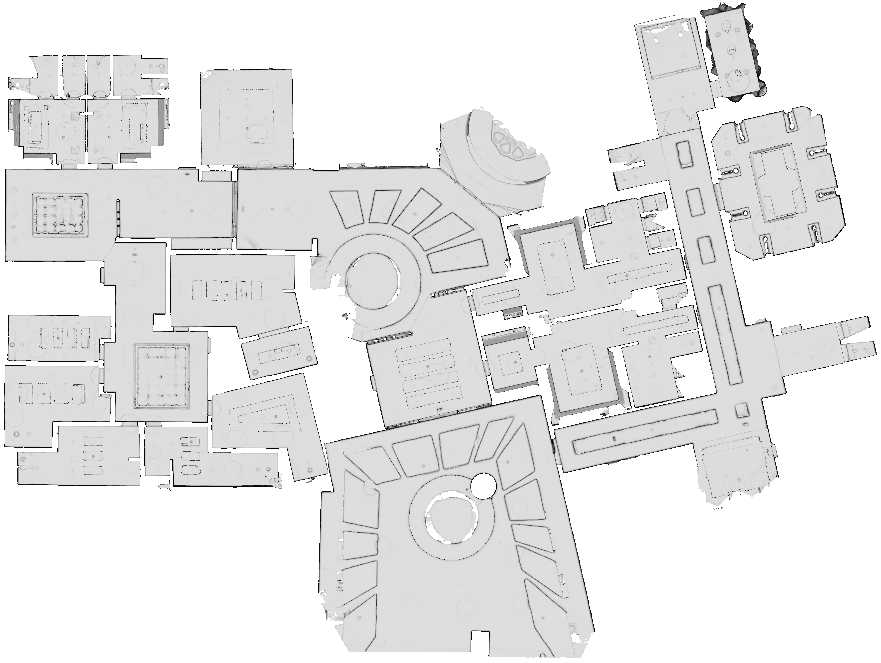}
        };
        \draw [draw=green, line width=1pt] (-5.05,-3.85) rectangle (5.05,3.85);
        \begin{scope}[rotate around={12.85:(0,0)}]
            \draw [draw=red, line width=1pt] (-5.2,-4.0) rectangle (4.85,4.15);    
        \end{scope}
    \end{tikzpicture}
    \caption{
        The green bounding box represents the horizontal alignment of the triangle mesh 'ur6pFq6Qu1A' from \autoref{fig:results_matterport3d_ur6pFq6Qu1A} as it is published in \citep{chang_et_al_2017} and used as ground truth pose for the evaluation results presented in  \autoref{tab:results}.
        The red bounding box on the other hand represents the horizontal alignment resulting from our presented approach.
    }
    \label{fig:discussion_matterport3d_ur6pFq6Qu1A}
\end{figure}

    This suggests that the proposed method consistently results in the same horizontal orientation for all 50 input rotations for both datasets.
    The respective resulting alignment however deviates from the assumed ground truth pose in the rotation around the vertical axis.
    This is further illustrated by \autoref{fig:discussion_matterport3d_ULsKaCPVFJR} and \autoref{fig:discussion_matterport3d_ur6pFq6Qu1A} where it is easily discernible that the depicted buildings again respectively contain two Manhattan World structures and that the evaluated alignment method consistently chooses the respective other Manhattan World structure that does not coincide with the ground truth pose.
    
    Arguably, it is disputable which of the two Manhattan World structures respectively present in the datasets is the 'correct' one as again in these two examples, both seem to encompass more or less the same fraction of the represented building environment and it is not readily discernable which is the dominant one.
    Nevertheless, our proposed method proofs to find a reasonable alignment with high accuracy in almost all cases with the only exception being the vertical alignment of the HoloLens triangle mesh 'Attic'.
    In all other cases where the resulting pose deviates from the ground truth pose, the resulting alignment is still reasonable in the sense that it corresponds to another Manhattan World structure inherent in the respective dataset that is readily identifiable by a human observer even if it may differ from the given ground truth pose corresponding to another alternative Manhattan World structure.
    
    Besides aligning an indoor mapping dataset with the dominant Manhattan World structure supported by the largest fraction of geometries, the proposed method can easily be augmented to identify all major Manhattan World structures along with the respective sets of associated geometries.
    Among other possible fields of application that will be briefly discussed in the following \autoref{sec:conclusions}, this allows for providing multiple possible alternatives for alignment to the user to choose from in cases where multiple major Manhattan World structures are present in the dataset at hand and it is not readily apparent which among these to use for alignment.
    \section{Conclusions}
\label{sec:conclusions}

In this work, we present a novel method for the automated pose normalization of indoor mapping data like point clouds and triangle meshes.
The aim of the proposed method is to align an indoor mapping point cloud or triangle mesh along the coordinate axes in a way that a chosen vertical axis points upwards with respect to the represented building structure, i.e. the chosen vertical axis is expected to be orthogonal to horizontal floor and ceiling surfaces.
Furthermore, a rotation around this vertical axis is to be determined in a way that aligns the two horizontal coordinate axes with the main direction of the dominant Manhattan World structure of the respective building geometry.
In case multiple Manhattan World systems are present in the data, the dominant structure supported by the largest fraction of geometries should determine the horizontal alignment.

For both fundamental steps of the proposed method - determining the correct orientation of the vertical axis and subsequently the correct horizontal rotation around this resulting vertical axis - a theoretical solution is presented.
As the proposed formulation of the problem at hand cannot be solved efficiently, an efficient approximate solution for a practical implementation is presented.
This encompasses discretizing the input data into a grid with fixed resolution while transforming it in a way that enables the problem to be solved by determining the largest peak within this grid of fixed size and finally refining the resulting coarse result by resorting to the original input data in the vicinity of the detected peak.
A CPU-parallelized implementation of the proposed method along with the code for the automated evaluation procedure will be made available to the public upon acceptance for publication.

The proposed method is quantitatively evaluated on a range of different indoor mapping point clouds and triangle meshes that are publicly available.
The presented results show, that the approach is overall able to consistently produce correct poses for the considered datasets for different input rotations with high accuracy.
Furthermore, cases where high deviations with respect to the given ground truth pose occur are presented and discussed. 

Concerning potential for future research, it has already been mentioned that the proposed method offers the possibility to not only identify the dominant Manhattan World structure along with the associated geometries in an indoor mapping dataset, but also to detect multiple Manhattan World structures that are sufficiently supported by geometries.
Besides enabling to present multiple reasonable alternatives for alignment to choose from, this could potentially also be used in the context of automated indoor reconstruction.
In particular, knowing the major Manhattan World structures and their associated geometries could be beneficial for abstracting and idealizing indoor surfaces, i.e. reconstructing suitable surfaces as planes that perfectly conform the Manhattan World assumption.
In addition, automatically detecting the involved Manhattan World structures in a building may also be of interest in the context of automatically analyzing the architectural structure of buildings \citep{ahmed_et_al_2011, roth_et_al_2020}.

Furthermore, the presented methodology could possibly also be used in the context of Simultaneous Localization and Mapping (SLAM) in indoor environments and indoor mapping in general.
Here, identifying Manhattan World structures during the mapping process (or in post processing if the individual indoor mapping geometries have associated timestamps to reconstruct the sequence of acquisition) could potentially be used to correct or reduce drift effects by applying the assumption that building structures that apparently seem to deviate only slightly from an ideal Manhattan World system are to be corrected according to the Manhattan World assumption \citep{peasley_et_al_2012,saurer_et_al_2012,straub_et_al_2015, straub_et_al_2018,yazdanpour_et_al_2019,liu_meng_2020}.
    \bibliographystyle{apalike}
    \bibliography{literature.bib}
\end{document}